\documentclass[sigconf,nonacm]{acmart}
\usepackage{bm}
\usepackage{amsfonts}
\usepackage{textcomp}
\newcommand{\mengqi}[1]{{\textcolor{black}{#1}}}

\usepackage{url}

\usepackage{amsthm}
\usepackage{multirow}
\usepackage{makecell}
\usepackage{subcaption}

\usepackage{listings}
\newsavebox{\promptdesignbox}

\usepackage[utf8]{inputenc}
\usepackage{colortbl}
\usepackage[linesnumbered,ruled,vlined]{algorithm2e}

\usepackage{pifont}
\newcommand{\cmark}{\ding{51}} 
\newcommand{\xmark}{\ding{55}} 

\usepackage{enumitem}

\newtheorem{definition}{\textbf{Definition}}

\newtheorem{example}{\textbf{Example}}

\newtheorem{lemma}{\textbf{Lemma}}
\newtheorem{remark}{Remark}

\newcommand{\TreeTool}{TreeED}
\newcommand{\ForestTool}{ForestED}
\newcommand{\FMED}{FM\_ED}
\newcommand{\ZeroED}{ZeroED}

\makeatletter
\patchcmd{\proof}
  {\topsep}
  {3pt \@plus 0.1ex \@minus 0.1ex}
  {}{}
\patchcmd{\endproof}
  {\topsep}
  {3pt \@plus 0.1ex \@minus 0.1ex}
  {}{}
\makeatother

\setcopyright{none}
\begin{document}

\title{Ensembling LLM-Induced Decision Trees for Explainable and Robust Error Detection}
\author{Mengqi Wang}
    \orcid{0009-0008-2900-7413}
    \affiliation{%
      \institution{University of New South Wales}
      \city{Sydney}
      \state{NSW}
      \country{Australia}
    }
    \email{mengqi.wang4@unsw.edu.au} 
    
\author{Jianwei Wang}
    \orcid{0009-0000-7887-4179}
    
    \affiliation{%
      \institution{University of New South Wales}
      \city{Sydney}
      \state{NSW}
      \country{Australia}
    }
    \email{jianwei.wang1@unsw.edu.au}
    \authornote{Jianwei Wang is the corresponding author. }
    
\author{Qing Liu}
    \orcid{0000-0001-7895-9551}

    \affiliation{%
      \institution{Data61, CSIRO}
      \city{Sandy Bay}
      \state{Tasmania}
      \country{Australia}
    }
 \email{q.liu@data61.csiro.au}
    
\author{Xiwei Xu}
  \orcid{0000-0002-2273-1862}
  \affiliation{%
      \institution{Data61, CSIRO}
        \city{Eveleigh}
        \state{NSW}
        \country{Australia}
      }
  \email{xiwei.xu@data61.csiro.au}

\author{Zhenchang Xing}
  \orcid{0000-0001-7663-1421}
  \affiliation{%
      \institution{Data61, CSIRO}
        \city{Canberra}
        \state{ACT}
        \country{Australia}
      }
  \email{zhenchang.xing@data61.csiro.au}

\author{Liming Zhu}
\orcid{0000-0001-5839-3765}
  \affiliation{%
      \institution{Data61, CSIRO}
        \city{Sydney}
        \state{NSW}
        \country{Australia}
      }
  \email{liming.zhu@data61.csiro.au}

\author{Michael Bain}
    \orcid{0000-0002-4309-6511}
    \affiliation{%
      \institution{University of New South Wales}
      \city{Sydney}
      \state{NSW}
      \country{Australia}
    }
    \email{m.bain@unsw.edu.au}

\author{Wenjie Zhang}
    \orcid{0000-0001-6572-2600}
    \affiliation{%
      \institution{University of New South Wales}
      \city{Sydney}
      \state{NSW}
      \country{Australia}
    }
    \email{wenjie.zhang@unsw.edu.au}

\renewcommand{\shortauthors}{Mengqi Wang et al.}

\begin{abstract}
Error detection (ED), which aims to identify incorrect or inconsistent cell values in tabular data, is important for ensuring data quality.
Recent state-of-the-art ED methods leverage the pre-trained knowledge and semantic capability embedded in large language models (LLMs) to directly label whether a cell is erroneous.
\mengqi{However, this \textit{LLM-as-a-labeler} pipeline (1) outputs raw detection results through an implicit, black-box process that lacks traceability, offering little explicit justification for why a particular prediction is made, and (2) relies on a single-pass inference that is inherently susceptible to model stochasticity, resulting in inconsistent detections and a lack of robustness across varying contexts.}
To address these limitations, we propose an \textit{LLM-as-an-inducer} framework that uses an LLM to induce a decision tree for ED (termed \TreeTool{}) and further ensembles multiple such trees for consensus detection (termed \ForestTool{}), thereby improving explainability and robustness.
Specifically, based on prompts derived from data context, decision tree specifications and output requirements, \TreeTool{} queries the LLM to induce the decision tree skeleton, whose root‑to‑leaf decision paths specify the stepwise procedure for evaluating a given sample.
Each tree contains three types of nodes: (1) rule nodes that perform simple validation checks (e.g., format or range), (2) Graph Neural Network (GNN) nodes that capture complex patterns (e.g., functional dependencies), and (3) leaf nodes that output the final decision types (error or clean).
Furthermore, \ForestTool{} employs uncertainty sampling to obtain multiple informative row subsets, constructing a decision tree for each subset using \TreeTool{}.
It then leverages an Expectation-Maximization-based algorithm that jointly estimates tree reliability and optimizes the consensus ED prediction.
Experiments demonstrate that our methods are accurate, explainable and robust, achieving an average F1-score improvement of 16.1\% over the best baseline. 
\end{abstract}

    \maketitle
    \newcommand\kddavailabilityurl{https://github.com/T-Lab/ForestED}
\label{code}
\ifdefempty{\kddavailabilityurl}{}{
\begingroup\small\noindent\raggedright\textbf{KDD Availability Link:}\\
The source code of this paper has been made publicly available at \url{\kddavailabilityurl}.
\endgroup
}

\begin{figure}
  \centering
  \includegraphics[width=0.96\linewidth]{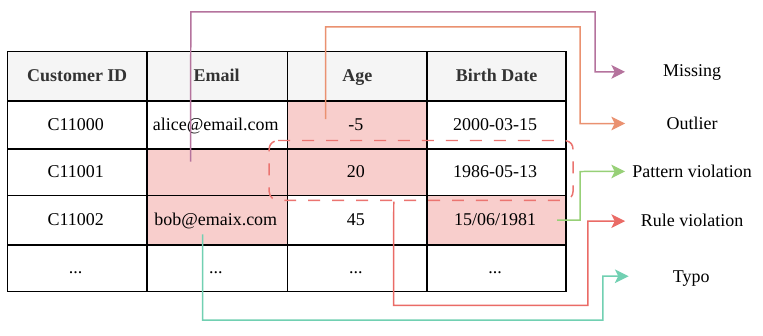}
\caption{An illustrated example of common data errors.}
  \label{fig:example}
\end{figure}

\section{Introduction}
\label{sec:Introduction}

Data quality is crucial for supporting reliable data analytics and data-driven decision-making. 
Poor data quality can distort analytical results, hinder operational processes, and ultimately lead to negative downstream impact~\cite{gupta2021data,wang2024missing,yu2024makes,zhu2024tassess}.
Unfortunately, real-world datasets often suffer from low quality due to issues like inconsistent data collection, leading to errors such as missing values, outliers, typos, and rule violations~\cite{fan2008conditional,chu2016data,ge2022hybrid}.
To address this issue, data cleaning~\cite{li2021cleanml,mahdavi2021semi,wu2025zero} is widely studied to substantially improve data quality through a two-phase process: (1) error detection (ED), which identifies incorrect or inconsistent cell values in tabular data, and (2) data repair, which corrects the identified errors. 
In this paper, we study ED, which is a critical step for effective data repair and downstream data preparation.

\begin{example}
    Common data error patterns are illustrated in Figure~\ref{fig:example}, including 1) The missing value error occurs where the \textit{Email} field is blank; 
    2) The outlier error appears in the \textit{Age} column where a negative value (–5) is invalid for a person's age; 
    3) a pattern violation occurs in the Birth Date column, where ``15/06/1981'' does not follow the expected ``YYYY-MM-DD'' format;
    4) a rule violation is shown between Age and Birth Date, where the reported age is inconsistent with the birth date;
    5) A typo occurs in the malformed email ``bob@emaix.com'', where the domain contains a misspelling.
\end{example}

Given the importance and practical value of ED, a set of methods has been developed~\cite{huang2018auto,heidari2019holodetect,pham2021spade}. Traditional ED methods for tabular data are mainly statistical-based~\cite{pit2016outlier,krishnan2016activeclean} and rule-based~\cite{ebaid2013nadeef,mahdavi2019raha,heidari2019holodetect}. 
Statistical-based approaches detect potentially erroneous values by modeling value distributions or correlations and identifying entries that deviate from expected patterns, but they can be sensitive to distributional variability and may incorrectly flag rare but legitimate values. 
On the other hand, rule-based approaches rely on rules, such as integrity constraints, patterns, functional dependencies (FDs), or knowledge bases, to identify violations. However, they require domain expertise and effort to construct and maintain such rules.

With the development of large language models (LLMs) for data-centric tasks~\cite{huang2024alchemist, wang2024human,long2024llms,ren2025few,wang2025onllm}, recent studies have increasingly explored their potential for data quality management~\cite{li2024llm,zhang2023jellyfish,parciak2024schema,zhang2025data,bendinelli2025exploring,wang2025match}. 
Building on these advances, LLM-based ED methods have emerged. They exploit the pretrained knowledge and semantic understanding capabilities of LLMs~\cite{tan2025hydra} to determine whether table cells are clean or erroneous, either by labeling each cell or by labeling a sampled subset of cells.
The first line of work, with \FMED{}~\cite{narayan2022can} as a representative, constructs a dedicated prompt for each cell in the table and uses the LLM to classify it. 
The second line of work, with \ZeroED{}~\cite{ni2025zeroed} as a representative, first employs the LLM to label a selected subset of cells. It then trains a multi-layer perceptron (MLP) classifier using these labels, and the classifier is applied to determine whether a cell is clean or erroneous.

\noindent \textbf{Motivations.} 
Although \FMED{} and \ZeroED{}, which follow the \textit{LLM-as-a-labeler} pipeline, achieve state-of-the-art (SOTA) performance for ED, they still face two notable limitations.

Firstly, their detection results lack explainability. As shown in Table~\ref{tab:ed_comparison}, they rely on black-box components where the LLM functions as an opaque labeler that directly outputs a final decision without providing a traceable rationale. Similarly, the MLP classifier in \ZeroED{} acts as a black-box neural model.
As a result, users cannot pinpoint the reasons that lead to the
predicted outcomes~\cite{petch2022opening}. 
This lack of explainability makes it difficult for experts to verify detected errors, impedes root cause diagnosis and undermines trust in automated data systems.

Secondly, the detection results lack robustness due to their reliance on a single-pass inference process. They task the LLM with making a final judgment in one execution, leaving the results highly susceptible to inherent stochasticity, where probabilistic decoding strategies and GPU-level variations cause identical prompts to yield inconsistent labels~\cite{jang2025rcscore}. Furthermore, this single-pass vulnerability is compounded by the MLP classifier in \ZeroED{}, which is highly sensitive to the quality of the labels it is trained on, including their correctness, diversity, and coverage. In the context of low-quality input data, such single-pass detections fail to provide the robustness required for practical data cleaning applications.

\begin{table}[t]
\centering
\caption{Comparison of ED methods.}
\label{tab:ed_comparison}
\resizebox{0.98\columnwidth}{!}{
\begin{tabular}{lcccc}
\toprule
\textbf{Approach} 
& \textbf{Label Free?} 
& \textbf{Backbone} 
& \textbf{LLM Role} 
& \textbf{Explainability} \\
\midrule

dBoost~\cite{pit2016outlier} 
& \cmark 
& Statistical Model 
& -- 
& \xmark  \\

Raha~\cite{mahdavi2019raha} 
& \xmark 
& Classifier 
& -- 
&  \xmark  \\

FM\_ED~\cite{narayan2022can} 
& \cmark  
& LLM 
& Labeler 
& \xmark  \\

ZeroED~\cite{ni2025zeroed} 
& \cmark  
& LLM + MLP
& Labeler 
& \xmark  \\

\midrule
\TreeTool{} (Ours) 
& \cmark  
& LLM + Decision Tree 
& Inducer 
& \cmark \\

\ForestTool{} (Ours) 
& \cmark  
& LLM + Decision Tree  
& Inducer 
& \cmark \\
\bottomrule
\end{tabular}
}
\vspace{2mm}
\end{table}

\noindent\textbf{Challenges.} To design an accurate, explainable and robust ED method, two challenges remain:

\textit{Challenge I: How to design an accurate and explainable ED method for noisy tabular data with complex relationships?}
The existence of diverse error patterns and complex table dependencies complicates the problem. A direct approach is to prompt LLMs to produce natural-language reasoning traces, such as Chain-of-Thought (CoT)~\cite{wei2022chain}, thereby providing an explanation for each error judgment. However, such self-generated explanations are often unfaithful or post-hoc rationalizations of the model decisions~\cite{yao2022react, matton2025walk, tan2025paths}, limiting their reliability. Another promising approach is to use LLMs to generate explicit logic rules for ED that are extracted from sampled data. This method offers a more structured form of explainability for the detection results. Nevertheless, simple rules struggle to represent diverse error patterns and capture complex relational structures among tables.

\textit{Challenge II: How to design a robust ED method for large-scale tabular data?}
The limited context windows of LLMs and the complex distributions of large real-world tabular data make robust ED particularly challenging. A naive approach is to prompt the LLM with the entire table, which can enhance robustness by leveraging global context. However, real-world datasets may contain tens or hundreds of thousands of tuples, far exceeding the input limits of most current LLMs. 
A more practical alternative, as in \ZeroED{}, is to label a subset of the table and train a downstream classifier on these partial labels. Nevertheless, a single detector struggles to characterize the intricate distributions and suffers from the stochasticity of LLM-based labeling.
Another line of work, exemplified by methods such as Raha~\cite{mahdavi2019raha}, trains a classifier to aggregate outputs from multiple detectors.
Yet these aggregation strategies rely on ground-truth labels, which are hard to obtain in real-world scenarios.

\noindent \textbf{Our approaches.}
Driven by the aforementioned challenges, we propose an \textit{LLM-as-an-inducer} framework that leverages LLMs to induce decision trees composed of different node types to handle diverse error patterns and errors of varying complexity (termed \TreeTool{}). We further ensemble multiple such trees by applying an Expectation–Maximization (EM)–based procedure to obtain consensus detection results (termed \ForestTool{}). In this way, we enhance the accuracy, explainability and robustness of ED in tabular data.

In \TreeTool{}, we use an LLM to induce a decision tree for ED (see case study in Figure~\ref{fig:case_study} for detailed illustration).
Specifically, we provide the LLM with data context, decision tree specification, and output requirements, which ground the LLM in the structure of the table and guide it toward producing meaningful decision logic. 
To capture diverse error patterns and complexities, we guide the LLM to induce a decision tree skeleton comprising three different nodes, including rule nodes, Graph Neural Network (GNN) nodes, and leaf nodes (representing error or clean states). 
This design effectively combines neural models and symbolic rules for enhanced performance.
Rule nodes encode simple symbolic checks such as format, range, and domain validation. 
In GNN nodes, we model the table as a bipartite graph and approach ED as a link classification task that leverages global context for superior performance under complex relational error patterns, such as cross-attribute correlations and functional dependencies. It can effectively capture constraints that simple rules cannot express. 
Moreover, each GNN node is dedicated to a specialized function, such as verifying the dependencies of a specific attribute, thereby ensuring explainability.
During prediction, each cell is evaluated through the tree. At every node of the tree, the corresponding rule or GNN check determines the next branch to follow, and the process continues until a leaf node is reached. This results in an explainable decision path.

To further enhance the robustness of the framework, \ForestTool{} is proposed. It consists of two components: (1) generating multiple decision trees through a multi-pass inference process and (2) ensembling their ED results.
We generate multiple decision trees by providing \TreeTool{} with distinct subsets of informative rows, selected via an uncertainty sampling strategy.
To aggregate the consensus results of these trees without labeled data, we propose an EM-based algorithm. 
The algorithm models both the unknown consensus predictions and the tree-specific reliabilities (i.e., error rates) as latent variables, which are iteratively estimated through EM.
Specifically, in the E-step, given the current tree reliability matrix and the prediction prior (i.e., the probability of each cell being erroneous), it updates the posterior consensus predictions. In the M-step, it updates the tree reliability matrix and the prediction prior via maximum likelihood estimation based on the estimated consensus predictions.
This iterative refinement continues until convergence, producing a consensus ED result that is more robust than a single tree.

\noindent \textbf{Contributions.} The main contributions are summarized as:
\begin{itemize}

    \item We propose an \textit{LLM-as-an-inducer} framework for error detection in tabular data, consisting of \TreeTool{} and \ForestTool{}.
    
    \item We introduce \TreeTool{}, which leverages an LLM to induce a decision tree that combines rule nodes for simple validation checks with GNN nodes for complex relational error patterns, enabling accurate and explainable ED.
    
    \item We introduce \ForestTool{}, which ensembles multiple \TreeTool{}-derived predictions using an EM–based consensus procedure to obtain reliable and robust ED results.
    \item We conduct extensive experiments across diverse datasets and evaluation settings, demonstrating that our framework outperforms existing baselines in detecting multiple error types, while substantially improving explainability and robustness, yielding an average F1-score improvement of 16.1\% over the strongest baseline.
\end{itemize}

\section{Problem Statement}
\label{sec:Preliminary}

Let $\mathbf{D}=\{\mathbf{T}_1,\ldots,\mathbf{T}_N\}$ be a dirty tabular dataset with $N$ tuples. 
The dataset contains $M$ attributes, denoted by $A_1,\ldots,A_M$. 
For each tuple $\mathbf{T}_i$, we use $\mathbf{D}[i,j]$ to represent the value of attribute $A_j$ in that tuple. 
Let $\mathbf{D}^{*}$ denote the corresponding clean (ground-truth) dataset. 
Tabular data can contain a wide variety of errors arising during data collection, entry, or integration. 
Common error types include missing values, typos, pattern violations, outliers, and rule violations~\cite{mahdavi2019raha, nashaat2021tabreformer, abdelaal2024saged, ni2025zeroed, chen2025minimum}. 
Missing values occur when certain fields are left empty, while typos result from manual entry mistakes such as misspellings or extra characters, both of which can often be identified from individual cell contents. 
Pattern violations occur when values deviate from expected syntactic formats (e.g., inconsistent date formats), whereas outliers refer to values outside the typical statistical or semantic range of an attribute. 
Rule violations capture cross-attribute inconsistencies, such as mismatched hierarchical relationships or logically impossible combinations, and often require understanding dependencies across multiple attributes or tuples.
Here, we give a formal definition of tabular error detection. The details about symbols used in this paper with descriptions are summarized in Appendix~\ref{app:symbol}.

\begin{definition}[Tabular Error Detection]
Given a dirty tabular dataset $\mathbf{D}$ with $N$ tuples and $M$ attributes, and the corresponding clean dataset $\mathbf{D}^{*}$, the ground-truth binary error matrix $\mathbf{Y} \in \{0,1\}^{N \times M}$ is defined as
\[
\mathbf{Y}[i,j] =
\begin{cases}
1, & \text{if } \mathbf{D}[i,j] \neq \mathbf{D}^{*}[i,j],\\
0, & \text{otherwise}.
\end{cases}
\]
The task of tabular error detection is to estimate a predicted error matrix $\hat{\mathbf{Y}} \in \{0,1\}^{N \times M}$ that approximates $\mathbf{Y}$.
\end{definition}

\begin{figure*}[t]
  \centering
  \includegraphics[width=0.98\linewidth]{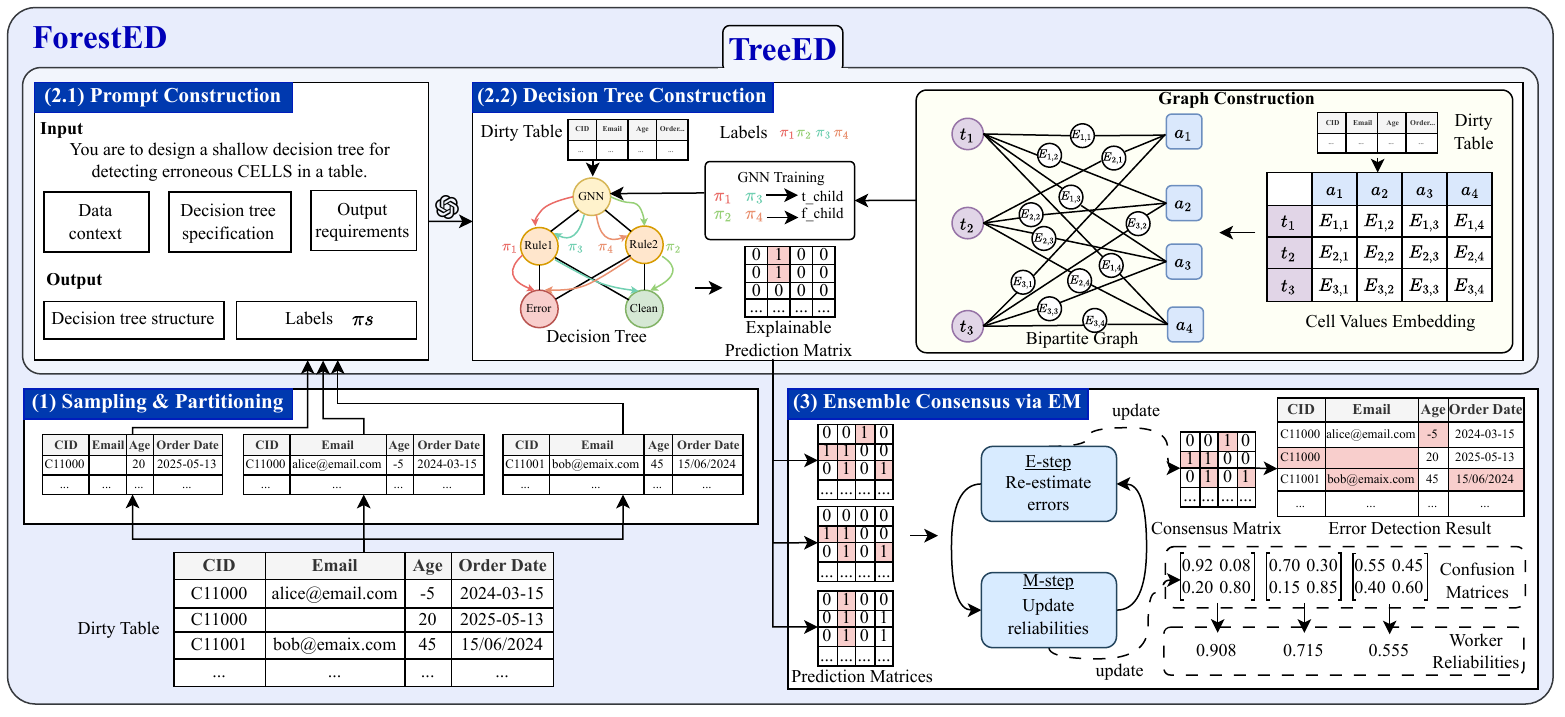}
  \caption{Framework overview of \TreeTool{} and \ForestTool{}. \TreeTool{} induces an LLM-generated decision tree with rule and GNN nodes to infer cell-level error labels via root-to-leaf execution, while \ForestTool{} ensembles multiple trees induced from sampled subsets and applies an EM-based consensus to aggregate their predictions into reliable final results.}
  \label{fig:frameworks_overview}
  \vspace{-4mm}
\end{figure*}

\section{\TreeTool{} Method}
\label{sec:tree_ed}
LLMs exhibit strong reasoning capabilities based on their prior knowledge, enabling them to induce complex symbolic structures, such as decision trees, from data and specifications~\cite{knauer2025oh, carrasco2025zero, kopanja2025cortex, xiong2024gptree}.
Leveraging this capability, \TreeTool{} prompts an LLM to generate an explainable, executable error detector in the form of a decision tree for tabular data and use it to perform error prediction.
The workflow of \TreeTool{} consists of two stages:  
(1) \textbf{prompt construction}, where an LLM is guided with structured context to generate a decision tree skeleton; and
(2) \textbf{decision tree construction}, where the induced tree skeleton is instantiated as an executable decision tree and then applied to the original table to identify erroneous cells. Algorithm of \TreeTool{} can be found in Appendix~\ref{sec:appendix_algorithms}.

\subsection{Prompt Construction} \label{subsec:prompt}

\mengqi{As shown in Figure~\ref{fig:frameworks_overview},} the prompt for constructing a decision tree in \TreeTool{} consists of three key components: data context, decision tree specification, and output requirements.  
An example prompt is provided in Appendix~\ref{sec:appendix_prompt}. A concrete example of an induced decision tree is provided in the case study in Section~\ref{sec:case_study} (see Figure~\ref{fig:case_study}).

\noindent\textbf{Data context}.
The data context provides essential background information for the LLM, consisting of two parts: (i) a data profile helping the LLM understand its structure and semantics without populating all tuples; and (ii) a set of representative sample rows illustrating diverse data patterns (selected by uncertainty sampling detailed in Section~\ref{sec:uncertain_sampling}). Specifically, the data profile is generated from the original dataset to provide dataset-level information that helps the LLM formulate relevant rules, following the idea of~\cite{fathollahzadeh2025catdb}. 
For each attribute $A_j$ in the schema, the data profile records its data type, distinctness, missing ratio, inclusion dependencies, inter-column similarity, correlation scores, and summary statistics (detailed in Appendix~\ref{sec:appendix_prompt}).

\noindent\textbf{Decision tree specification}.  
The decision tree specification defines both the structure and the functional roles of different node types. Since real-world tabular datasets exhibit error patterns of varying complexity, from simple syntactic errors to multi-attribute relational violations, the tree includes three complementary node types: symbolic rule nodes for simple local checks, learning-based GNN nodes for complex global reasoning, and leaf nodes for final classification. 
By modeling the table as a bipartite graph and propagating information over the bipartite graph, GNNs aggregate contextual signals from related rows and attributes, enabling the detection of inconsistencies that arise only from global structure. 
Finally, leaf nodes act as terminal decision points that output whether the examined cell is erroneous or clean.
\begin{remark}
Although GNN nodes are not fully interpretable, as their internal decision logic remains black-box, their roles within the decision tree are path-level explainable. Specifically, (1) each node has an LLM-induced descriptive message indicating its intended function (e.g., checking an FD, as shown in Figure~\ref{fig:case_study}); and (2) the node also specifies the attributes involved in each relational dependency and is dedicated to the specific checking function.
\end{remark}

\noindent\textbf{Output requirements}.  
To induce a valid and executable decision tree from the LLM, we restrict the output format to two core components: a decision tree structure \texttt{tree\_structure} and a set of labeled sample records \texttt{labels}. These two components jointly specify how the tree is constructed and how it behaves on example inputs, ensuring an unambiguous induced tree.

\subsection{Decision Tree Construction} \label{subsec:decision_tree_construction}
Given the two LLM outputs, the \texttt{tree\_structure} component and the \texttt{labels} component, \TreeTool{} constructs an executable decision tree $\mathcal{T}$ with a set of nodes $\mathcal{U}(\mathcal{T})$ that integrates both rule nodes $\mathcal{U}_{rule}$ and GNN nodes $\mathcal{U}_{gnn}$. 
Since the rule nodes directly incorporate the LLM-generated code without further processing, we next detail the training process for the GNN nodes in three steps: (1) label preparation, (2) graph construction, and (3) GNN training.

\noindent \textbf{Step 1: label preparation.}
To train GNN nodes, we extract supervision labels from the \texttt{labels} component, which records the decision paths that sampled cells follow from the root to a leaf in the decision tree.
For each sampled cell $\mathbf{D}[i,j]$, its root-to-leaf decision path in the tree $\mathcal{T}$ is represented as
\[
\pi(i,j) = \big[(u_0, b_0), (u_1, b_1), \ldots, (u_K, b_K)\big],
\]
where $u_k$ denotes the $k$-th node encountered along the path, and $b_k \in \{0,1\}$ indicates which outgoing branch is taken at that node (\texttt{0} for the false branch and \texttt{1} for the true branch).  
The values of $b_k$ only denote which branch is chosen at each decision point and do not directly indicate whether the cell is an error.

\begin{example}
Consider the decision path for cell $\mathbf{D}[i,j]$:\\
$\pi(i,j) =
[(\texttt{check\_country\_code},0),\,
 (\texttt{gnn\_fd\_location},1),\,\\
 (\texttt{check\_location\_gdp\_range},0),\,
 (\texttt{leaf\_ok},0)]$.
This means the cell first reaches the rule node \texttt{check\_country\_code} and selects the false branch (e.g., the format check does not trigger), then reaches the GNN node \texttt{gnn\_fd\_location} and selects the true branch (e.g., the relational pattern is satisfied), then selects the false branch at \texttt{check\_location\_gdp\_range}, and ultimately arrives at \texttt{leaf\_ok}, indicating that the cell is labeled as clean.
\end{example}
From these decision paths, we derive supervision labels for training the GNN nodes. 
For each GNN node $u_k$ that appears in the path $\pi(i,j)$, where $(u_k, b_k) \in \pi(i,j)$, the corresponding cell $\mathbf{D}[i,j]$ is assigned a supervision label $l_{i,j,k} = b_k$, indicating which branch the cell follows at node $u_k$. 
By collecting all cells whose root-to-leaf path includes node $u_k$ and pairing them with their associated labels, we then obtain the training set $\mathcal{L}_{u_k}$ corresponding to $u_k$.

\noindent \textbf{Step 2: graph construction.}
We employ a graph-based representation to support relational reasoning, as many tabular errors arise from dependencies that span multiple rows or multiple attributes and therefore cannot be captured by local, cell-level checks alone. 
To model these interactions, we represent the table as a bipartite graph motivated by ~\cite{you2020handling}.
In the constructed bipartite graph, tuple nodes and attribute nodes form two disjoint sets, and each cell corresponds to an edge connecting its associated tuple and attribute. 
This bipartite design naturally reflects the two-dimensional structure of tabular data and enables information propagation across both tuple and attribute dimensions, making it particularly effective for capturing error patterns that require cross-row or cross-column reasoning (e.g., functional dependencies). Formally, each GNN node operates on a bipartite graph:
\[
\mathcal{G} = (\mathcal{V}_t \cup \mathcal{V}_a, \mathcal{E}),
\]
where $\mathcal{V}_t$ and $\mathcal{V}_a$ denote tuple and attribute nodes, respectively.
Tuple node $v_t^i \in \mathcal{V}_t$ corresponds to the $i$-th row in the original table, while attribute node $v_a^j \in \mathcal{V}_a$ corresponds to the $j$-th column.  
An edge ${e_{{v_t^i, v_a^j}}} \!\in\! \mathcal{E}$ between node $v_t^i$ and node $v_a^j$ represents a table cell $\mathbf{D}[i,j]$ and is associated with a feature vector:
\[
\mathbf{h}_{e_{{v_t^i, v_a^j}}} = \phi(\mathbf{D}[i,j]),
\]
where $\phi(\cdot)$ is a hash-based embedding function that encodes the cell value into a numerical vector.

\noindent \textbf{Step 3: GNN training.}
For each GNN node $u_k$ in the induced decision tree, we train an independent GNN on its constructed bipartite graph $\mathcal{G}_{u_k}$ with supervision labels $\mathcal{L}_{u_k}$. 
Following GRAPE~\cite{you2020handling}, we extend the GraphSAGE architecture~\cite{hamilton2017inductive} by incorporating explicit edge embeddings.  
For node initialization, tuple nodes are assigned an $m$-dimensional constant vector, whereas attribute nodes are given an $m$-dimensional one-hot vector that aims to uniquely identify each attribute:
\[
\mathbf{h}_{v}^{(0)} =
\begin{cases}
\mathbf{1}, & v \in \mathcal{V}_t, \\[4pt]
\text{ONEHOT}(v), & v \in \mathcal{V}_a.
\end{cases}
\]
At each GNN layer $p$, the neighborhood representation of a node $v$ is obtained by forming a message for every neighbor $v' \in \mathcal{N}(v)$ via concatenating the neighbor’s previous-layer embedding $\mathbf{h}_{v'}^{(p-1)}$ with the corresponding edge embedding $\mathbf{h}_{e_{v',v}}$, passing it through a nonlinear transformation, and aggregating all such messages: \[ \mathbf{h}_{\mathcal{N}(v)}^{(p)} = \text{AGG}\!\left\{ \sigma\!\left( \text{CONCAT}\!\left( \mathbf{h}_{v'}^{(p-1)},\, \mathbf{h}_{e_{v',v}} \right) \right) \;\middle|\; v' \in \mathcal{N}(v) \right\}. \] 
Finally, the node embedding is updated by combining its previous-layer embedding of $\mathbf{h}_v^{(p-1)}$ with its corresponding aggregated neighborhood representation $\mathbf{h}_{\mathcal{N}(v)}^{(p)}$: 
\[ \mathbf{h}_{v}^{(p)} = \sigma\!\left( \mathbf{W}^{(p)} \cdot \text{CONCAT}\!\left(\mathbf{h}_v^{(p-1)},\, \mathbf{h}_{\mathcal{N}(v)}^{(p)}\right) \right). \]
Here, $\text{AGG}(\cdot)$ denotes a permutation-invariant mean aggregator, 
$\mathbf{W}^{(p)}$ is the trainable weight matrix at propagation layer $p$, 
and $\sigma(\cdot)$ is a nonlinear activation function RELU.
After two propagation layers, we obtain the final embeddings for all tuple and attribute nodes.
For each cell $\mathbf{D}[i,j]$, we perform binary routing prediction by feeding the concatenated tuple node and attribute node embeddings into a two-layer MLP:
\[
\hat{q}_{i,j,k}
=
\text{MLP}\!\left(
\mathbf{h}_{v_{t}^i}^{(2)} \,\|\, \mathbf{h}_{v_{a}^j}^{(2)}
\right),
\]
which outputs the probability of routing the cell to the true or false branch at GNN node $u_k$.
The model is trained using binary cross-entropy loss:
\[
Loss_k = -\!\sum_{(i,j)}\! \Big[l_{i,j,k}\log \hat{q}_{i,j,k} + (1-l_{i,j,k})\log (1-\hat{q}_{i,j,k})\Big].\]
where $l_{i,j,k} \in \{0,1\}$ is the branch supervision label derived from the LLM-induced decision path.
After training, each GNN node predicts the branch decision for a cell by applying its learned GNN embeddings to the optimized MLP classifier.

Once the decision tree $\hat{\mathcal{T}}$ is constructed, each table cell is evaluated by traversing a root-to-leaf path that applies rule or GNN checks at internal nodes, with the reached leaf providing the binary error label, yielding a cell-level ED matrix $\hat{\mathbf{Y}}$.

\section{\ForestTool{} Method}
\label{sec:forest_ed}

While \TreeTool{} demonstrates that an LLM can effectively serve as a decision tree inducer for ED, its performance may vary due to the stochastic nature of LLMs.
To address this, \ForestTool{} applies uncertainty sampling to select an informative subset of rows and partitions it into smaller segments, with each segment inducing its own decision tree to form a diverse forest of detectors.
We then ensemble their predictions using an EM-based consensus procedure inspired by~\cite{dawid1979maximum, wang2025ensemble}, which jointly infers the latent true labels and each tree-specific reliability to achieve a consensus ED result (illustrated in Figure~\ref{fig:frameworks_overview}). Algorithm of \ForestTool{} can be found in Appendix~\ref{sec:appendix_algorithms}.

\subsection{Sampling and Partitioning}\label{sec:uncertain_sampling}

\noindent\textbf{Motivation.}
Annotating every cell in a large dataset with an LLM is prohibitively expensive, so we focus on labeling only a small but highly informative subset of rows. Rather than relying on random sampling, we target tuples that lie in irregular, ambiguous, or boundary-case regions of the data distribution, as these are more likely to reveal informative error patterns~\cite{krishnan2016activeclean}.
Prioritizing such uncertain regions reduces annotation cost and exposes the LLM to diverse hard cases, improving generalization.

To identify informative tuples, we adopt a Gaussian Process (GP)–based uncertainty sampling strategy, a standard technique for retrieving valuable subsets from large-scale datasets~\cite{zhao2021learned}. 
For each tuple $i$, we construct a feature vector $\mathbf{x}_i \in \mathbb{R}^d$ by standardizing numeric attributes, one-hot encoding categorical attributes, and applying PCA  to reduce dimension.  
A GP places a prior over a latent function $g(\cdot)$ that models similarity between tuples, using an RBF kernel
\[
\mathcal{K}(\mathbf{x}, \mathbf{x}') = \exp\!\left(-\frac{\lVert \mathbf{x} - \mathbf{x}' \rVert_2^2}{2\chi^2}\right),
\]
where $\chi$ is the kernel length-scale controlling how quickly similarity decays with distance.  
Fitting the GP on a small subset with a dummy target yields a posterior distribution, from which we obtain a predictive variance for each tuple $\mathrm{Var}\!\left[g(\mathbf{x}_i)\right]$
where larger variance indicates greater uncertainty~\cite{lewis1994heterogeneous}.
We then select the top-$s$ tuples with the highest variances, where $s = \min\!\left(100,\, \lceil \rho N \rceil\right)$
and $\rho$ is the sampling ratio.  

Specifically, the selected samples $\mathcal{S}$ are divided into $R$ disjoint partitions $\{\mathcal{S}^{(1)},\mathcal{S}^{(2)},\ldots,\mathcal{S}^{(R)}\}$, each serving as the input sample for LLM to induce a decision tree $\hat{\mathcal{T}}^{(r)}$.  
By construction, $\hat{\mathcal{T}}^{(r)}$ models error patterns from its corresponding partition $\mathcal{S}^{(r)}$, while the full collection $\{\hat{\mathcal{T}}^{(r)}\}_{r=1}^R$ forms a decision forest, and the corresponding prediction set $\{\hat{\mathbf{Y}}^{(r)}\}_{r=1}^R$ is later aggregated by EM into a unified consensus result.

\begin{lemma}[Informativeness of Uncertainty-Based Sampling] \label{lem:uncertainty_sampling}
Let $\mathcal{D}$ be a tabular dataset and $\mathcal{T}$ be a decision tree induced by an LLM to approximate the true error-detection boundary $\mathcal{B}$. Let $\mathcal{S}_u$ and $\mathcal{S}_r$ be subsets of $\mathcal{D}$ of size $k$, where $\mathcal{S}_u$ is selected via uncertainty sampling and $\mathcal{S}_r$ via random sampling. In limited data contexts (small $k$), the expected Information Gain $\mathbb{E}[I]$ provided by $\mathcal{S}_u$ is strictly greater than that of $\mathcal{S}_r$:
\begin{equation}
    \mathbb{E}[I(\mathcal{S}_u)] > \mathbb{E}[I(\mathcal{S}_r)].
\end{equation}
\end{lemma}
Consequently, samples selected via uncertainty sampling provide more informative supervision signals for model induction.
As training on more informative samples leads to improved model quality and generalization~\cite{lewis1994heterogeneous}, decision trees induced from $\mathcal{S}_u$ tend to better approximate the underlying error patterns than those induced from randomly sampled subsets.

Proof of Lemma~\ref{lem:uncertainty_sampling} can be found in Appendix~\ref{proof:uncertainty_sampling}.

\subsection{Ensemble Consensus via EM Algorithm} \label{subsec:em_consensus}

\textbf{Motivation.}
When ensembling predictions from multiple decision trees, simple strategies such as majority voting ignore that different trees may have very different reliability due to variations in sampled rows, LLM behavior, and induced structures.
To overcome this, we adopt an EM-based consensus procedure that treats both the true cell labels and tree-specific reliabilities as latent variables. By iteratively estimating these reliabilities and updating the consensus labels to approximate true cell labels, EM down-weights unreliable trees, reinforces consistent signals, and produces a stable ensemble.

\noindent\textbf{Latent variables and model parameters.}
For each cell $\mathbf{D}[i,j]$, we model its ground-truth error label as a latent random variable $\mathbf{Y}_{i,j} \in \{0,1\}$.  
Each induced decision tree $\hat{\mathcal{T}}^{(r)}$ provides an observed prediction $\hat{\mathbf{Y}}^{(r)}_{i,j} \in \{0,1\}$.  
The probability that the latent label is $y$ given all tree predictions is denoted by the posterior:
\[
\gamma_{i,j}(y) 
    = p(\mathbf{Y}_{i,j}=y \mid \hat{\mathbf{Y}}^{(1)}_{i,j},\dots,\hat{\mathbf{Y}}^{(R)}_{i,j}), 
\]
and the prior distribution over error labels is written as:
\[
\eta_y = p(\mathbf{Y}_{i,j}=y), \qquad y\in\{0,1\}.
\]
To capture the reliability of each tree, we associate $\hat{\mathcal{T}}^{(r)}$ with a $2\times 2$ latent confusion matrix
\[
\Theta^{(r)} =
\begin{pmatrix}
\theta^{(r)}_{00} & \theta^{(r)}_{01} \\
\theta^{(r)}_{10} & \theta^{(r)}_{11}
\end{pmatrix},
\qquad
\theta^{(r)}_{y,\hat{y}}
= p(\hat{\mathbf{Y}}^{(r)}=\hat{y} \mid \mathbf{Y}=y),
\]
where each $\theta^{(r)}_{y,\hat{y}}$ in $\theta^{(r)}_{y,\hat{y}}$ characterizes how likely tree $\hat{\mathcal{T}}^{(r)}$ is to output $\hat{y}$ when the true label is $y$.
The EM algorithm iteratively alternates between the E-step and M-step until convergence, and then performs a consensus prediction to obtain the final ED result.

\noindent\textbf{Step 1: E-step.}
The E-step updates the posterior of the true cell labels given the reliability matrix of each tree and the prior distribution computed in the M-step. 
Specifically, it applies Bayes' rule to compute the posterior responsibility: 
\[
\gamma_{i,j}(y)
=
\frac{
    \eta_y
    \prod_{r=1}^R
    \theta^{(r)}_{\,y,\,\hat{\mathbf{Y}}^{(r)}_{i,j}}
}{
    \sum_{y'\in\{0,1\}}
    \eta_{y'}
    \prod_{r=1}^R
    \theta^{(r)}_{\,y',\,\hat{\mathbf{Y}}^{(r)}_{i,j}}
}.
\]
The likelihood term reflects how each tree tends to behave when the true label is $y$, while the prior $\eta_y$ accounts for the overall error rate in the dataset.  
Multiplying them captures how strongly the tree predictions support label $y$ in the context of expected error frequency.
Normalizing over both labels yields a valid posterior distribution that reflects how likely each cell is erroneous.

\noindent\textit{Initialization in the first iteration.}
The first E-step initializes the posterior responsibilities directly from the 
\TreeTool{} predictions. For each cell, the initial posterior is obtained by normalizing the aggregated predictions from all trees, yielding a data-driven estimate of how likely the cell is to be clean or erroneous:
\[
\gamma^{(0)}_{i,j}(y)
=
\frac{\sum_{r} \mathbf{1}\{\hat{\mathbf{Y}}^{(r)}_{i,j}=y\}}
     {\sum_{y'\in\{0,1\}}\sum_{r}\mathbf{1}\{\hat{\mathbf{Y}}^{(r)}_{i,j}=y'\}}.
\]

\noindent\textbf{Step 2: M-step.}
In the M-step, given the posterior of true cell labels, we compute the reliability matrix of each tree and the prior distribution over error labels. 
For each tree $\hat{\mathcal{T}}^{(r)}$, the $(y,\hat{y})$ entry of its confusion matrix is computed as
\[
\theta^{(r)}_{y,\hat{y}}
=
\frac{
    \sum_{i,j: \,\hat{\mathbf{Y}}^{(r)}_{i,j} = \hat{y}}
    \gamma_{i,j}(y)
}{
    \sum_{i,j} \gamma_{i,j}(y)
},
\qquad y,\hat{y}\in\{0,1\}, 
\]
where the numerator is the expected number of cells with true label $y$ that tree $\hat{\mathcal{T}}^{(r)}$ predicts as $\hat{y}$, and the denominator is the expected number of cells with true label $y$.
The prior distribution is approximated using the empirical mean of the posteriors:
\[
\eta_y = \frac{1}{NM}\sum_{i,j}\gamma_{i,j}(y).
\]
As established in Lemma~\ref{lemma1} and its proof in Appendix~\ref{proof}, the update rules in both the E-step and M-step monotonically improve the consensus quality of the ensemble.
\begin{lemma}[Monotonicity of EM Updates]\label{lemma1}
Let $o_{i,j}(y)$ be any variational distribution over the latent label 
$\mathbf{Y}_{i,j}$, and define the evidence lower bound (ELBO) as
\[
\mathcal{F}(o,\Theta)
=
\sum_{i,j}\sum_{y}
o_{i,j}(y)\Big[
\log \eta_y
+
\sum_{r=1}^R 
\log \theta^{(r)}_{y,\hat{\mathbf{Y}}^{(r)}_{i,j}}
-
\log o_{i,j}(y)
\Big].
\]
For an ensemble of $R$ decision trees producing predictions 
$\hat{\mathbf{Y}}^{(r)}_{i,j}$, maximizing $\mathcal{F}$ with 
respect to either the variational posteriors $o$ (E-step) or the 
reliability matrices $\Theta$ (M-step) yields a 
monotonic increase in the marginal log-likelihood 
$\log p(\hat{\mathbf{Y}}\mid\Theta)$.
Consequently, each EM update improves the consensus quality of the ensemble.
\end{lemma}
Proof of Lemma~\ref{lemma1} can be found in Appendix~\ref{proof}.

\noindent\textbf{Step 3: consensus prediction.}
After convergence, the posterior matrix 
$\Gamma = [\gamma_{i,j}(y)]$ 
encodes the final consensus estimate for all cells.  
The unified ED result is obtained by taking the MAP label:
\[
\hat{\mathbf{Y}}_{i,j}
    = \arg\max_{y\in\{0,1\}} \gamma_{i,j}(y).
\]

Additional analysis of explainability and time complexity is provided in 
Appendix~\ref{sec:appendix_analysis}.

\section{Experimental Evaluation}
\label{sec:Experiment}

\subsection{Experimental Setup}
\noindent\textbf{Datasets}. We evaluate our approach on seven datasets (Table~\ref{tab:datasets}) following existing ED methods~\cite{ni2025zeroed, mahdavi2019raha, narayan2022can}. 
The datasets contain diverse error types, including missing values (MV), pattern violations (PV), typos (T), outliers (O), and rule violations (RV).

\noindent\textbf{Baselines.}  
We evaluate \ForestTool{} against a range of SOTA ED methods for tabular data, including statistics-based approaches (dBoost~\cite{pit2016outlier}), rule-based systems (Raha~\cite{mahdavi2019raha}), and LLM-based methods (FM\_ED~\cite{narayan2022can}, ZeroED~\cite{ni2025zeroed}). We also report the performance of our \TreeTool{}, which uses one LLM-induced decision tree for ED. 

\begin{table}[t]
\centering
\small
\caption{Dataset statistics used for evaluation.}
\label{tab:datasets}
\setlength{\tabcolsep}{4.5pt}
\begin{tabular}{|c|c|c|c|c|}
\hline
\cellcolor{lightgray}\textbf{Dataset} &
\cellcolor{lightgray}\textbf{Tuples} &
\cellcolor{lightgray}\textbf{Attributes} &
\cellcolor{lightgray}\textbf{Errors(\%)} &
\cellcolor{lightgray}\textbf{Error Types} \\ \hline
Rayyan       & 1{,}000    & 11 & 29.19 & PV, T, O, RV\\ \hline
Hospital     & 1{,}000    & 17 & 4.82 &  MV, PV, T, O, RV\\ \hline
Flights      & 2{,}376    & 6  & 34.51&  MV, PV, T, O, RV\\ \hline
Beers        & 2{,}410    & 9 & 12.98 & MV, PV, T, O, RV\\ \hline
Billionaire  & 2{,}614    & 22 & 9.16  & MV, PV, T, O, RV \\ \hline
Movies       & 7{,}390    & 17 & 4.97  & MV, PV, T, O \\ \hline
Tax          & 200{,}000  & 15 & 4.05  & MV, PV, T, O, RV \\ \hline
\end{tabular}
\end{table}

\begin{table*}[t]
\centering
\small
\caption{
Overall Precision (Prec), Recall (Rec), and F1-score (F1).
GPT-5 is used as the default LLM.
Following ZeroED, a SOTA method of ED, we provide two labeled samples per dataset for baselines requiring supervision, while all others are evaluated in a zero-shot setting.
All remaining hyperparameters follow the default settings of the respective methods.
}

\label{tab:main_results_sorted}
\setlength{\tabcolsep}{2pt}
\begin{tabular}{
|p{1.6cm}<{\centering}|
p{0.58cm}<{\centering}p{0.58cm}<{\centering}p{0.58cm}<{\centering}| 
p{0.58cm}<{\centering}p{0.58cm}<{\centering}p{0.58cm}<{\centering}| 
p{0.58cm}<{\centering}p{0.58cm}<{\centering}p{0.58cm}<{\centering}| 
p{0.58cm}<{\centering}p{0.58cm}<{\centering}p{0.58cm}<{\centering}| 
p{0.58cm}<{\centering}p{0.58cm}<{\centering}p{0.58cm}<{\centering}| 
p{0.58cm}<{\centering}p{0.58cm}<{\centering}p{0.58cm}<{\centering}| 
p{0.58cm}<{\centering}p{0.58cm}<{\centering}p{0.58cm}<{\centering}| 
}
\hline
\multirow{2}{*}{\textbf{Methods}} &
\multicolumn{3}{c|}{\textbf{Rayyan}} &
\multicolumn{3}{c|}{\textbf{Hospital}} &
\multicolumn{3}{c|}{\textbf{Flights}} &
\multicolumn{3}{c|}{\textbf{Beers}} &
\multicolumn{3}{c|}{\textbf{Billionaire}} &
\multicolumn{3}{c|}{\textbf{Movies}} &
\multicolumn{3}{c|}{\textbf{Tax}} \\ 
\cline{2-22}
 & Prec & Rec & F1 & Prec & Rec & F1 & Prec & Rec & F1 & Prec & Rec & F1 & Prec & Rec & F1 & Prec & Rec & F1 & Prec & Rec & F1 \\ 
\hline
dBoost & 0.343 & 0.361 & 0.352 & 0.227 & 0.074 & 0.112 & 0.524 & 0.388 & 0.446 & 0.363 & 0.147 & 0.210 & 0.484 & 0.142 & 0.220 & 0.145 & 0.092 & 0.112 & 0.577 & 0.274 & 0.371 \\ 
Raha   & 0.610 & 0.540 & 0.570 & 0.460 & 0.090 & 0.140 & 0.660 & 0.320 & 0.430 & 0.760 & 0.640 & 0.680 & 0.170 & 0.100 & 0.130 & 0.230 & 0.270 & 0.250 & 0.400 & 0.640 & 0.487 \\ 
\hline
FM\_ED  & 0.500 & 0.500 & 0.500 & 0.132 & \textbf{0.844} & 0.224 & 0.537 & 0.302 & 0.325 & 0.192 &0.246 & 0.195 & 0.432& 0.498 &0.459 & 0.108 & 0.639 & 0.138 & 0.027 & 0.115 & 0.006 \\ 
ZeroED & 0.751 & \textbf{0.631} & 0.686 & 0.449 & 0.650 & 0.531 & 0.493 & \textbf{0.720} & 0.585 & 0.548 & \textbf{0.957} &0.678 & 0.384 & \textbf{0.746} & 0.455 & 0.700 & \textbf{0.696} & 0.698 & 0.399 & 0.689 & 0.505 \\
\hline
TreeED (avg) & 0.995 & 0.550 & \textbf{0.709} & 0.638 & 0.416 & 0.472 & \textbf{1.000} & 0.535 & 0.697 & 0.741 & 0.928 & 0.782 & 0.644 & 0.369 & 0.358 & 0.700 & 0.576 & 0.583 & \textbf{1.000} & \textbf{0.721} & \textbf{0.838} \\ 
ForestED      & \textbf{0.998} & 0.544 & 0.704 & \textbf{0.844} & 0.685 & \textbf{0.756} & \textbf{1.000} & 0.562 & \textbf{0.720} & \textbf{0.974} & 0.922 & \textbf{0.947} & \textbf{0.698} & 0.560 & \textbf{0.566} & \textbf{0.934} & 0.606 & \textbf{0.735} & \textbf{1.000} & \textbf{0.721} & \textbf{0.838}\\ 
\hline
\end{tabular}
\vspace{-4mm}
\end{table*}

\begin{table*}[t]
\centering
\small
\caption{Ablation study on components of ForestED across datasets.}
\label{tab:ablation_sorted}
\setlength{\tabcolsep}{2pt}
\begin{tabular}{
|p{1.6cm}<{\centering}|
p{0.58cm}<{\centering}p{0.58cm}<{\centering}p{0.58cm}<{\centering}| 
p{0.58cm}<{\centering}p{0.58cm}<{\centering}p{0.58cm}<{\centering}| 
p{0.58cm}<{\centering}p{0.58cm}<{\centering}p{0.58cm}<{\centering}| 
p{0.58cm}<{\centering}p{0.58cm}<{\centering}p{0.58cm}<{\centering}| 
p{0.58cm}<{\centering}p{0.58cm}<{\centering}p{0.58cm}<{\centering}| 
p{0.58cm}<{\centering}p{0.58cm}<{\centering}p{0.58cm}<{\centering}| 
p{0.58cm}<{\centering}p{0.58cm}<{\centering}p{0.58cm}<{\centering}| 
}
\hline
\multirow{2}{*}{\textbf{Ablation}} &
\multicolumn{3}{c|}{\textbf{Rayyan}} &
\multicolumn{3}{c|}{\textbf{Hospital}} &
\multicolumn{3}{c|}{\textbf{Flights}} &
\multicolumn{3}{c|}{\textbf{Beers}} &
\multicolumn{3}{c|}{\textbf{Billionaire}} &
\multicolumn{3}{c|}{\textbf{Movies}} &
\multicolumn{3}{c|}{\textbf{Tax}} \\ 
\cline{2-22}
 & Prec & Rec & F1 & Prec & Rec & F1 & Prec & Rec & F1 & Prec & Rec & F1 & Prec & Rec & F1 & Prec & Rec & F1 & Prec & Rec & F1 \\ 
\hline

w/o. gp.
& 0.944 & 0.614 & \textbf{0.744 } 
& 0.941 & 0.468 & 0.625  
& \textbf{1.000} & \textbf{0.563} & \textbf{0.720} 
& 0.970 & 0.070 & 0.130  
& \textbf{0.871} & 0.495 & \textbf{0.631}  
& 0.623 & 0.600 & 0.611  
& \textbf{1.000} & \textbf{0.724} & \textbf{0.840} \\

w/o. gnn
& 0.476 & \textbf{0.919} & 0.627  
& \textbf{1.000} & 0.398 & 0.569  
& \textbf{1.000} & 0.470 & 0.639  
& 0.975 & 0.706 & 0.819  
& 0.212 & \textbf{0.806} & 0.336  
& 0.104 & \textbf{0.720} & 0.182  
& \textbf{1.000} & \textbf{0.724} & \textbf{0.840} \\

w/o. ens.
& \textbf{0.998} & 0.544 & 0.704 
& 0.818 & 0.390 & 0.529  
& \textbf{1.000} & \textbf{0.563} & \textbf{0.720} 
& \textbf{0.989} & \textbf{0.951} & \textbf{0.970 } 
& 0.817 & 0.303 & 0.442  
& 0.932 & 0.598 & 0.721
& \textbf{1.000} &  0.721 & 0.838\\

ForestED 
& \textbf{0.998} & 0.544 & 0.704 
& 0.844 & \textbf{0.685} & \textbf{0.756}  
& \textbf{1.000} & \textbf{0.563} & \textbf{0.720} 
& 0.974 & 0.922 & 0.947
& 0.698 & 0.560 & 0.566  
& \textbf{0.934} & 0.606 & \textbf{0.735}  
& \textbf{1.000} & 0.721 & 0.838\\
\hline
\end{tabular}
\vspace{-2mm}
\end{table*}

\noindent\textbf{Metrics.}
In this paper, we primarily use \textit{Precision}, \textit{Recall}, and \textit{F1-score}, widely adopted by existing ED studies~\cite{mahdavi2019raha, ni2025zeroed}.
For all three metrics, higher values indicate better ED performance.

\noindent\textbf{Implementation details.}  
We use \texttt{gpt-5} as the default LLM via the OpenAI API with temperature fixed to 1.0.  
For uncertainty-driven sampling, we set the sampling ratio to $\rho = 0.05$ with at most 100 sampled tuples, and each partition contains 10 tuples for decision tree induction. For \ZeroED{}, we use the same ratio without a cap, following the original paper.
Induced decision trees have 4--8 levels, and each GNN node uses a two-layer GraphSAGE encoder with hidden dimension 64, learning rate $10^{-2}$, and 2000 training epochs.  
For EM-based consensus, EM runs for up to 100 rounds or until $\Delta < 10^{-4}$.
As \ForestTool{} does not rely on manually defined constraints or domain knowledge, we provide only 2 labeled tuples per dataset for baselines requiring supervision, following~\cite{ni2025zeroed}. 
Due to the high token usage of LLM-based baselines, we randomly sample 50 rows for \FMED{} and 500 rows for \ZeroED{} for evaluation.
Experiments run on a workstation 
with an Intel(R) Xeon(R) Silver~4314 CPU~@~2.40\,GHz, an NVIDIA RTX~A5000 GPU, and 512\,GB RAM.

\subsection{Overall Evaluation}

\vspace{0.5mm}
\noindent\textbf{Exp-1: Overall detection performance.}
Table~\ref{tab:main_results_sorted} summarizes the performance across seven datasets.
Traditional baselines such as dBoost and Raha perform poorly, consistently yielding much lower F1-scores than other methods, particularly on challenging datasets such as \textit{Hospital}, \textit{Billionaire}, \textit{Movies}, and \textit{Tax}, where the error rate is below 10\%.
LLM-based baselines improve performance but still show limitations: \FMED{} is highly unstable across datasets and sometimes underperforms traditional methods, whereas \ZeroED{} is more consistent and generally strong.
\TreeTool{} performs strongly on many datasets, though its effectiveness varies with the induced tree structure and attribute characteristics.
On the \textit{Rayyan} dataset, its performance is relatively stronger, likely because the higher error rate makes error patterns more salient and allows a single induced decision tree to capture most errors effectively.
\ForestTool{} further improves consistency and achieves the best performance on nearly all datasets, offering greater robustness than \TreeTool{}.
In terms of F1-score, it achieves average absolute gains of 49.2\% over dBoost, 36.8\% over Raha, 48.8\% over \FMED{}, 16.1\% over \ZeroED{}, and 11.8\% over \TreeTool{}, demonstrating the effectiveness of reliability-aware EM consensus for robust and accurate ED.

\vspace{1mm}

\noindent\textbf{Exp-2: Ablation study.}
Table~\ref{tab:ablation_sorted} evaluates the contributions of three key components in \ForestTool{}: GP-based uncertainty sampling, GNN-based relational reasoning, and EM-based ensemble consensus.
Removing GP-based uncertainty sampling (\textit{w/o. gp.}), where uncertainty sampling is replaced by random sampling, results in a clear performance degradation, particularly in recall, indicating that random or non-informative sampling exposes the LLM to fewer boundary and hard cases during tree induction and leads to weaker rule coverage. Interestingly, on particular datasets, removing uncertainty sampling yields slight performance improvements, which we attribute to reduced sampling bias when the error distribution is relatively uniform or when uncertainty estimates are noisy due to limited feature expressiveness. Overall, removing uncertainty sampling leads to an average F1-score drop of approximately 14.5\%, confirming that uncertainty sampling plays a critical role in selecting informative tuples and improving the quality and generalization of LLM-induced decision trees.
Removing GNN nodes (\textit{w/o. gnn}), leaving the decision tree composed solely of rule nodes, consistently lowers recall and results in an average F1-score drop of about 18.2\%, indicating that rule-based checks alone cannot capture relational dependencies.
Removing EM consensus (\textit{w/o. ens.}), where EM-based aggregation is replaced by simple majority voting across trees, also degrades performance, with an average F1-score reduction of roughly 4.9\%, reflecting that individual trees induced from different partitions may be biased or unstable when used alone.
While the full \ForestTool{} model is not always the top performer on every metric for every dataset, it generally achieves the most balanced and stable results overall. 
This further demonstrates that uncertainty sampling selects more informative subsets, GNNs broaden structural coverage, and EM-based consensus over multi-pass LLM inference calibrates tree reliability and mitigates the impact of unreliable or hallucinated tree structures. 

\begin{figure}[t]
    \centering
    \setlength{\abovecaptionskip}{2pt}
    \setlength{\belowcaptionskip}{-2pt}

    \begin{subfigure}{0.23\textwidth}
        \centering
        \includegraphics[width=\linewidth]{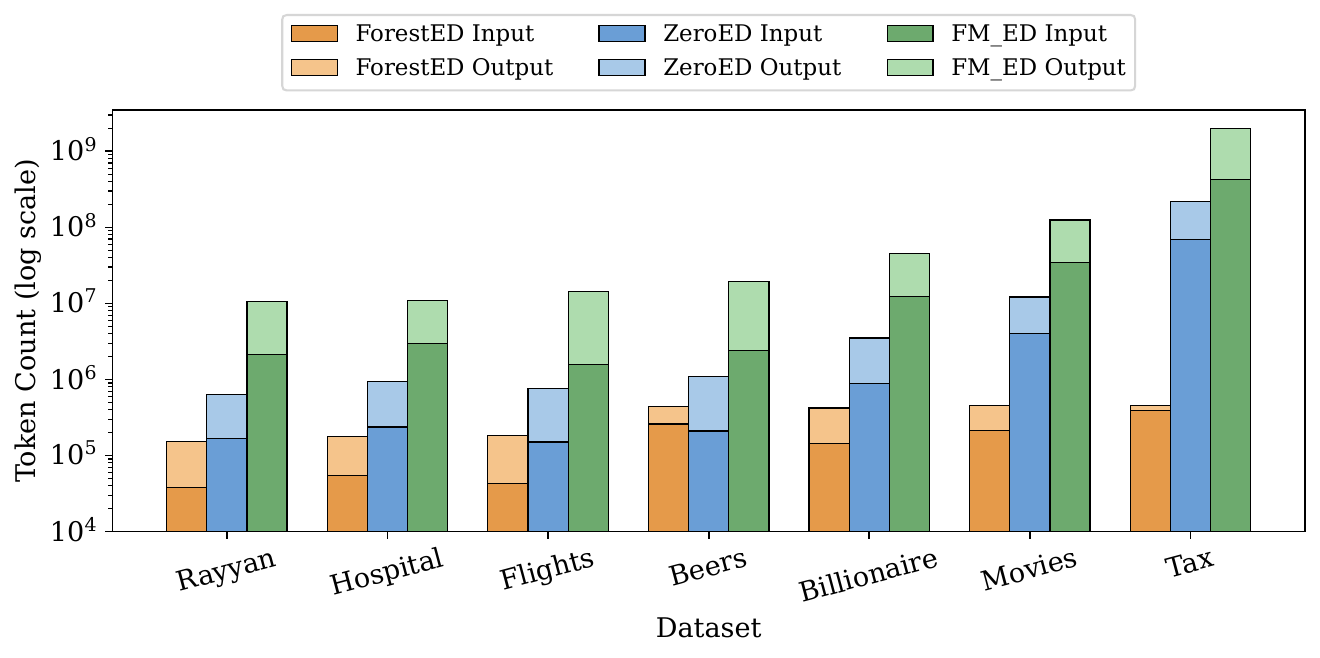}
        \caption{Token consumption}
        \label{fig:exp3_token_cost}
    \end{subfigure}
    \hspace{-2mm}
    \begin{subfigure}{0.23\textwidth}
        \centering
        \includegraphics[width=\linewidth]{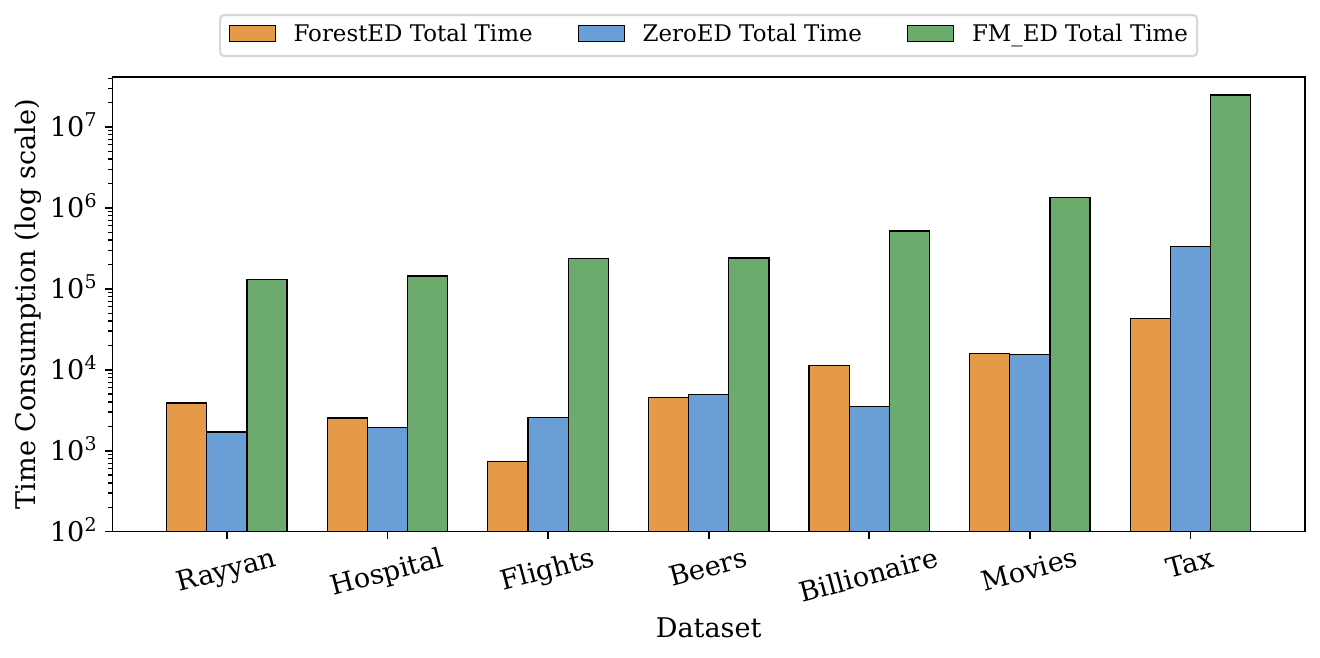}
        \caption{Running time}
        \label{fig:exp3_time_cost}
    \end{subfigure}

    \caption{Runtime and token cost across datasets.}
    \label{fig:exp3_cost}
\end{figure}
\vspace{1mm}
\noindent\textbf{Exp-3: Efficiency evaluation.}
This experiment compares the token consumption and runtime efficiency of \ForestTool{}, \FMED{}, and \ZeroED{} across multiple datasets.
As shown in Figure~\ref{fig:exp3_token_cost}, \ForestTool{} requires significantly fewer tokens than both \ZeroED{} and \FMED{}.
In terms of runtime, as shown in Figure~\ref{fig:exp3_time_cost}, \ForestTool{} is slightly slower on small datasets but becomes more efficient as the dataset size increases, whereas \ZeroED{} exhibits the opposite pattern.
Both methods, however, run much faster than \FMED{}.
Overall, \ForestTool{} achieves the best balance of token efficiency and time efficiency among SOTA LLM-based ED approaches.

\subsection{Robustness Evaluation}

\noindent\textbf{Exp-4: Mean and standard deviation over 3 trials.}
Table~\ref{tab:exp7_variance} reports the mean and standard deviation of precision, recall, and F1-score over three runs. Both LLM-related methods \FMED{} and \ZeroED{} show high instability, reflecting strong sensitivity to fluctuations in LLM-generated labels and classifier stochasticity. Rule-based method Raha is relatively more stable. Compared with \TreeTool{}, \ForestTool{} further reduces variation while improving the mean F1-score. Overall, \ForestTool{} achieves the highest accuracy and the smallest standard deviation in F1-score. Its average F1-score standard deviation is consistently the lowest across datasets, representing a 7× reduction compared with \FMED{} and a 10× reduction compared with \ZeroED{}. This substantial decrease indicates that multi-pass LLM inference, combined with EM-based ensemble consensus, improves overall robustness.

\begin{table}[t]
\centering
\caption{Mean and standard deviation of the metrics for Raha, \FMED{}, \ZeroED{}, \TreeTool{}, and \ForestTool{} over three trials.}
\label{tab:exp7_variance}

\resizebox{0.46\textwidth}{!}{
\begin{tabular}{|c|ccc|ccc|}
\hline
\multirow{2}{*}{\textbf{Methods}} &
\multicolumn{3}{c|}{\textbf{Beers}} &
\multicolumn{3}{c|}{\textbf{Billionaire}} \\ \cline{2-7}
 & Prec(std) & Rec(std) & F1(std) & Prec(std) & Rec(std) & F1(std) \\ \hline
Raha &
0.923(0.093) & 0.833(0.084)	 & 0.873(0.070)	 &
0.350(0.311) & 0.113(\textbf{0.078})	 & 0.150(0.085) \\
\FMED{} &
0.192 (0.147) & 0.246 (0.215) & 0.195 (0.141) &
0.432 (\textbf{0.131}) & 0.498 (0.195) & 0.459 (0.160) \\

\ZeroED{} &
0.548 (0.232) & \textbf{0.957} (\textbf{0.007}) & 0.678 (0.178) &
0.384 (0.257) & \textbf{0.746} (0.106) & 0.455 (0.245) \\

\TreeTool{} &
0.741 (0.296) & 0.928 (0.090) & 0.782 (0.206) &
0.644 (0.346) & 0.369 (0.230) & 0.358 (0.133) \\

\textbf{\ForestTool{} }&
\textbf{0.974} (\textbf{0.006}) & 0.922 (0.027) & \textbf{0.947} (\textbf{0.016}) &
\textbf{0.698} (0.238) & 0.560 (0.219) & \textbf{0.566} (\textbf{0.027}) \\ \hline
\end{tabular}
}
\end{table}

\begin{table}[t]
\centering
\caption{Recall comparison of ED methods across five error types on \textit{Hospital}.}
\label{tab:hospital3}
\begin{tabular}{|c|c|c|c|c|c|}
\hline
\textbf{Methods} & \textbf{MV} & \textbf{PV} & \textbf{T} & \textbf{O} & \textbf{RV} \\
\hline
Raha     & 0.573 & 0.271 & 0.026 & 0.000 & 0.281 \\
\FMED{}     & 0.600 & 0.833 & \textbf{0.900} & 0.429 & 0.000 \\
\ZeroED{}  & 0.782 & 0.692 & 0.545 & \textbf{0.611} & 0.740 \\
\textbf{ForestED} & \textbf{0.858} & \textbf{0.865} & 0.808 & \textbf{0.611} & \textbf{0.844} \\
\hline
\end{tabular}
\end{table}

\noindent\textbf{Exp-5: Varying error types.}
Table~\ref{tab:hospital3} compares the recall of different ED methods across error types on \textit{Hospital}.
Raha exhibits limited recall across most error categories, particularly for T and O errors, indicating that rule-based approaches struggle to capture complex or context-dependent error patterns.
\FMED{} performs unevenly across five error types, detecting typos well through pretrained linguistic knowledge but failing to identify rule violations due to insufficient access to table-level relational context.
ZeroED substantially improves average recall over Raha and \FMED{}, demonstrating the effectiveness of LLM-based semantic reasoning.
Furthermore, \ForestTool{} consistently achieves the highest recall across nearly all error types, with notable gains for PV and RV errors, highlighting its effectiveness in capturing both pattern-level inconsistencies and inter-column relational violations.

\begin{figure}[t]
    \centering
    \setlength{\abovecaptionskip}{2pt}
    \setlength{\belowcaptionskip}{-4pt}

    \begin{subfigure}{0.225\textwidth}
        \centering
        \includegraphics[width=\linewidth]{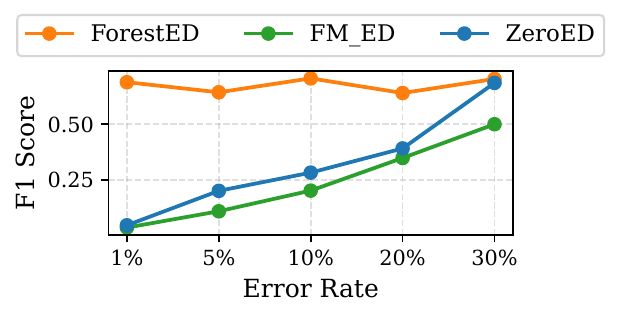}
        \caption{Rayyan}
    \end{subfigure}
    \hspace{-2mm}
    \begin{subfigure}{0.225\textwidth}
        \centering
        \includegraphics[width=\linewidth]{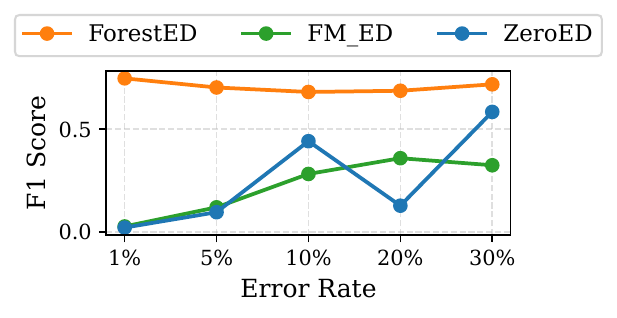}
        \caption{Flights}
    \end{subfigure}

    \caption{Effect of varying error rates.}
    \label{fig:exp6_error_rate}
\end{figure}

\noindent\textbf{Exp-6: Varying error rates.}
We evaluate the robustness of the three ED methods, \ForestTool{}, \ZeroED{}, and \FMED{}, under increasing error rates ranging from $1\%$ to $30\%$.  
As shown in Figure~\ref{fig:exp6_error_rate}, \ForestTool{} consistently achieves the highest F1-scores across all error-rate settings on both \textit{Rayyan} and \textit{Flights}.  
Its performance remains largely stable as the number of errors increases, exhibiting only minor fluctuations between the $1\%$ and $30\%$ settings.  
In contrast, \ZeroED{} and \FMED{} perform substantially worse at low error rates, although both methods improve as more error cells become available for learning.  
This behavior highlights their reliance on higher error rates to establish reliable decision boundaries. 
Overall, the results demonstrate that \ForestTool{} is markedly more resilient to varying degrees of errors, maintaining strong detection accuracy where \ZeroED{} and \FMED{} degrade, especially in low-error conditions.

\noindent \textbf{Additional hyperparameter and LLM backbones analysis.} We also conduct extensive experiments to evaluate the robustness of \ForestTool{}, including varying numbers of labels, LLMs, tree numbers, and tree depths, as detailed in Appendix~\ref{sec:appendix_experiment}. 
Results show that \ForestTool{} maintains stable performance under limited samples, benefits from additional samples with diminishing returns (Exp-9), exhibits significantly lower sensitivity to different LLM backbones than prior methods (Exp-10), achieves optimal accuracy with a moderate number of trees that balances diversity and stability (Exp-11), and performs best with medium-depth trees (Exp-12).

\subsection{Explainability Evaluation} \label{sec:explain_eval}
\begin{figure}[t]
    \centering
    \setlength{\abovecaptionskip}{2pt}

    \begin{subfigure}{0.155\textwidth}
        \centering
        \includegraphics[width=\linewidth]{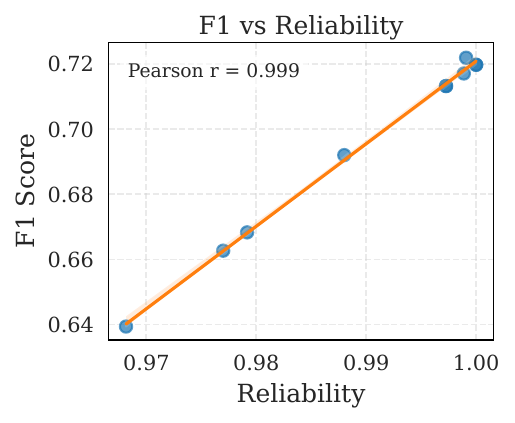}
        \caption{Flights}
    \end{subfigure}
    \hspace{-3mm}
    \begin{subfigure}{0.155\textwidth}
        \centering
        \includegraphics[width=\linewidth]{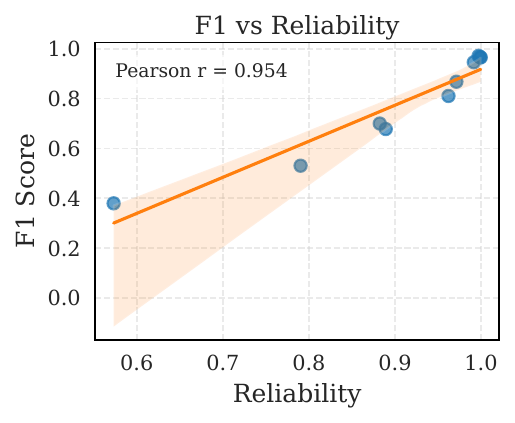}
        \caption{Beers}
    \end{subfigure}
    \hspace{-3mm}
    \begin{subfigure}{0.155\textwidth}
        \centering
        \includegraphics[width=\linewidth]{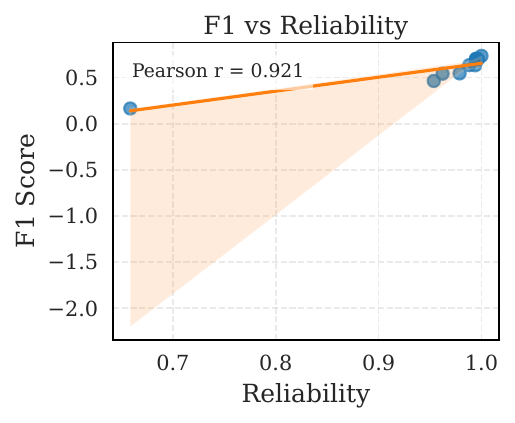}
        \caption{Movies}
    \end{subfigure}

    \caption{Correlation between F1 and ensemble reliability.}
    \label{fig:exp2_f1_reliability}
\end{figure}
\noindent\textbf{Exp-7: F1-score vs. ensemble reliability.}
We assess whether the EM-estimated reliability scores meaningfully reflect the predictive quality of individual partitions.
As shown in Figure~\ref{fig:exp2_f1_reliability}, F1-scores and reliability values exhibit a clear positive linear trend across datasets, with Pearson correlations above 0.7.
This indicates that partitions assigned higher reliability by the EM algorithm consistently achieve higher F1-scores, demonstrating that the consensus procedure accurately captures variations in partition quality.
Overall, these results show that EM reliability serves as an explainable and trustworthy indicator of prediction consistency within \ForestTool{}.

\begin{figure}[t]
  \centering
  \includegraphics[width=0.90\linewidth]{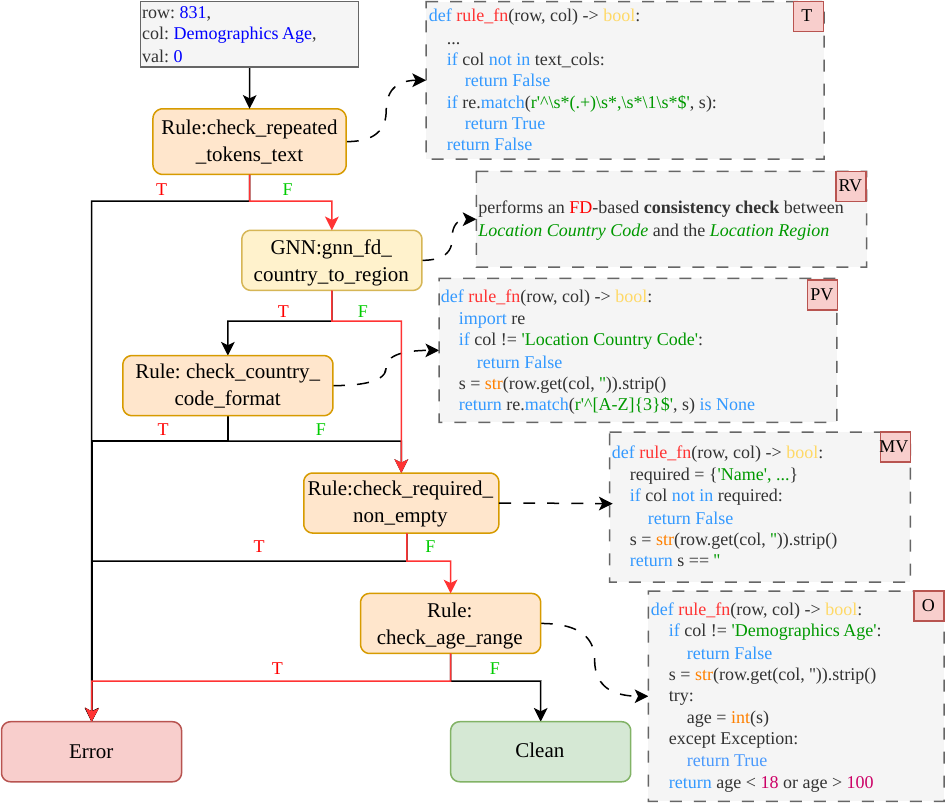}
\caption{An example of an induced decision tree.}
\label{fig:case_study}
\end{figure}

\noindent\textbf{Exp-8: Case study.} \label{sec:case_study}
We present a representative case to illustrate how an induced decision tree in \TreeTool{} detects different error types through explainable checks, as shown in Figure~\ref{fig:case_study}.
Specifically, we examine an erroneous cell from the \textit{Billionaire} dataset at row 831 and column \textit{Demographics Age}, where the observed value is ``0'' while the correct value is ``37''.
Since the inspected attribute is non-textual, the execution bypasses the root typo-checking rule and proceeds to the GNN node \texttt{gnn\_fd\_country\_to\_region}, which checks relational consistency between \texttt{Location Country Code} and \texttt{Location Region}.
As the age value does not violate this relational constraint, the execution continues to the next rule node \texttt{check\_required\_non\_empty}. As the cell is not empty, it is further routed to \texttt{check\_age\_range}.
This rule validates whether the age lies within a plausible range (18--100), identifies ``0'' as an outlier, and routes the tuple to \texttt{leaf\_error}.
Rule nodes dominate the induced tree in this case study, indicating that GNN nodes are used selectively for relational checks while most decision steps remain rule-based and directly traceable, thereby supporting path-level explainability.
In practice, ten independently induced trees produce split predictions for this cell, with five voting error and five voting clean; the EM-based consensus aggregates their reliabilities and assigns a final error probability of 99.99\%.
We also analyze the explainability and time complexity of \ForestTool{} in Appendix~\ref{sec:appendix_analysis}.

\vspace{-2mm}

\section{Related Work}
\label{sec:Relatedwork}

Error detection in relational data has evolved through four major paradigms.
Early approaches relied on integrity constraints, such as denial constraints, to detect logical inconsistencies, with work on scalable discovery (e.g., Hydra~\cite{bleifuss2017efficient}) and efficient violation detection (e.g.,  Katara~\cite{chu2015katara}, FACET~\cite{pena2021fast}).
Statistical methods later introduced corpus-driven notions of value compatibility; for example, dBoost~\cite{pit2016outlier} and Auto-Detect~\cite{huang2018auto} identify anomalies that violate global co-occurrence patterns rather than explicit rules.
A third line of work reframed error detection as a weakly supervised learning problem, including HoloDetect~\cite{heidari2019holodetect}, ActiveClean~\cite{krishnan2016activeclean}, and Raha~\cite{mahdavi2019raha}, which leverage limited labels or clean seeds to train data-driven detectors.
Most recently, foundation-model–based approaches such as FM\_ED~\cite{narayan2022can} and ZeroED~\cite{ni2025zeroed} enable zero-shot detection by prompting over serialized tuples, exploiting broad world knowledge to capture contextual inconsistencies beyond the reach of traditional ED methods.

\vspace{-2mm}
\section{Conclusion}
\label{sec:Conclusion}
We present an explainable and robust LLM-based framework for tabular ED under the paradigm of \textit{LLM-as-an-inducer}.  
Unlike prior methods that treat LLMs as black-box labelers or rely on downstream classifiers without exposing the underlying decision process, our approach introduces traceable decision structures together with ensemble consensus to improve both explainability and robustness.  
Specifically, \TreeTool{} leverages LLMs to induce decision trees composed of rule and GNN nodes, providing traceable decision paths for cell-level error detection, while \ForestTool{} enhances robustness by aggregating multiple induced trees through an EM-based consensus to consolidate consistent predictions and mitigate variance across runs.
Extensive experiments on diverse datasets demonstrate the effectiveness of our methods over existing ED methods.

\clearpage
\bibliographystyle{ACM-Reference-Format}
\bibliography{ref}  

\balance
\clearpage

\appendix

\section{Symbols and Descriptions}\label{app:symbol}
Table~\ref{tab:symbol} summarizes the key symbols and notations used throughout the paper. 

\begin{table}[t]
\centering 
\caption{Symbols and descriptions.}
\label{tab:symbol}
\begin{tabular}{|p{2.25cm}|p{5.55cm}|}
\hline
\cellcolor{lightgray}\textbf{Notation} & \cellcolor{lightgray}\textbf{Description} \\ \hline
$\mathbf{D}$ & Dirty dataset of size $N \times M$. \\ \hline
$\mathbf{D}^{*}$ & Clean (ground-truth) version of $\mathbf{D}$. \\ \hline
$\mathbf{T}_i$ / $t_i$ & $i$-th tuple / its index in $\mathbf{D}$, $i \in \{1,\ldots,N\}$. \\ \hline
$A_j$ / $a_j$ & $j$-th attribute / its index in $\mathbf{D}$, $j \in \{1,\ldots,M\}$. \\ \hline
 $\hat{\mathbf{Y}}$ & Estimated error matrix  $\hat{\mathbf{Y}} \in \{0,1\}^{N \times M}$. \\ \hline
$f_{\text{LLM}}(\cdot)$ & LLM inference function. \\ \hline
$\mathcal{G}=(\mathcal{V}_t\cup\mathcal{V}_a, \mathcal{E})$ 
& Bipartite graph over tuples and attributes.\\ \hline
$\mathcal{T}$ & A decision tree. \\ \hline
$\mathcal{U}(\mathcal{T})$:$\{u_1, \ldots, u_K\}$ & Node set of tree $\mathcal{T}$. \\ \hline
$\pi(i,j)$ & Root-to-leaf decision path of $\mathbf{D}[i,j]$ in $\mathcal{T}$. \\ \hline
$\mathcal{S}$; $\mathcal{S}^{(1)},\ldots,\mathcal{S}^{(R)}$ & Sample of $\mathbf{D}$ and its $R$ partitions. \\ \hline
$\Theta^{(r)}$ & Confusion matrix at the $r$-th tree. \\ \hline
$\hat{\mathbf{Y}}^{(r)}_{i,j}$ & Cell-level prediction by the $r$-th tree. \\ \hline
\end{tabular}
\end{table}

\section{Prompt} \label{sec:appendix_prompt}

An example of prompt is given in this box.

\begin{figure}[h]
\centering
\begingroup
\setlength{\fboxsep}{0pt}
\setlength{\fboxrule}{0.6pt}

\begin{lrbox}{\promptdesignbox}
\begin{minipage}{0.90\columnwidth}

\noindent\colorbox{blue!70!black}{%
  \makebox[\linewidth][l]{%
    \hspace{7pt}\textcolor{white}{\strut\textbf{Prompt Design}}%
  }%
}

\vspace{4pt}

\hspace{8pt}%
\begin{minipage}{\dimexpr\linewidth-16pt\relax}
\small
\raggedright
\setlength{\parindent}{0pt}
\setlength{\parskip}{2pt}

\textbf{1. Data context:} data profile, sample rows\\
\textbf{2. Decision tree specification:} The decision tree should be shallow (4--8 levels). Each node should be specialized, not overloaded with unrelated checks. There should be at least one \texttt{gnn} node whose children are not leaf nodes.
  \begin{itemize}[leftmargin=2.0em,labelsep=0.5em,itemsep=2pt,topsep=3pt,parsep=0pt]
\item \textbf{rule}: must provide executable Python code that performs simple rule checks, formatted as:
\begin{lstlisting}[language=Python, basicstyle=\ttfamily\footnotesize]
def rule_fn(tuple, attr) -> bool:
    # return True if the cell is erroneous,
    # False otherwise
\end{lstlisting}
  Each rule node should handle \textbf{only one type of check} 
  (e.g., numeric range, format, or cross-column consistency).  

  \item \textbf{gnn}: for complex relational checks
  (e.g., functional dependencies, conditional FDs, denial constraints), where the node name reflect the purpose (e.g., \texttt{gnn\_fd\_check}, \texttt{gnn\_cfd\_check}).  

  \item \textbf{leaf}: must have a \texttt{leaf\_value} of \texttt{true} or \texttt{false}.  
\end{itemize}
\textbf{3. Output requirements:}  
\begin{itemize}[leftmargin=2.0em,labelsep=0.5em,itemsep=2pt,topsep=3pt,parsep=0pt]
\item For each sampled row and every column, output one entry in \texttt{labels}. 
  Each entry must contain the \textbf{row\_id}, \textbf{column\_name},  \textbf{is\_error}, and the \textbf{full path of node\_ids from root to leaf}.
  \item The overall output must be a single JSON object with two fields:  
   \begin{itemize}[leftmargin=1.8em,labelsep=0.45em,itemsep=1pt,topsep=2pt,parsep=0pt]
     \item \texttt{"tree\_structure"}: list of \texttt{NodeSpec} objects.  
     \item \texttt{"labels"}: list of entries with  
           \texttt{row\_id}, \texttt{column}, \texttt{is\_error}, and \texttt{path}.  
   \end{itemize}
  \item {Example error types and output format}.  
\end{itemize}

\end{minipage}
\vspace{5pt}

\end{minipage}
\end{lrbox}

\fcolorbox{blue!70!black}{blue!5!white}{\usebox{\promptdesignbox}}

\endgroup
\label{box:prompt}
\end{figure}

For each attribute, its data profile includes: 
(1) \emph{data type}, such as numeric, categorical, or date, inferred from the value domain; 
(2) \emph{distinctness}, measured as the ratio of unique values to non-missing entries; 
(3) \emph{missing ratio}, i.e., the fraction of tuples where $A_j$ is empty; 
(4) \emph{inclusion dependencies}, indicating whether the values of $A_j$ are contained in another attribute; 
(5) \emph{inter-column similarity}, computed by cosine similarity between 300-dimensional column embeddings that encode value distributions using histograms, summary statistics, hashed buckets for numeric attributes, and hashed bag-of-tokens representations for non-numeric attributes; 
(6) \emph{correlation scores}, such as Pearson correlation or mutual information for numeric or categorical pairs; 
(7) \emph{representative values}, consisting of a small set of frequent samples; and 
(8) \emph{summary statistics}, including mean, median, standard deviation, and min--max values for numeric attributes.

For output requirements, \texttt{"tree\_structure"} defines the decision tree skeleton as a list of node objects. Each node includes a \texttt{node\_id} (unique identifier), a \texttt{type} (node type), and a \texttt{name} (semantic label). Rule nodes contain an executable \texttt{code} field specifying a Python function for simple checks, along with \texttt{true\_child} and \texttt{false\_child} pointers that determine branching behavior. GNN nodes represent relational checks and include only skeleton descriptions without explicit code. Leaf nodes specify a \texttt{leaf\_value} indicating whether the final decision is \texttt{true} (error) or \texttt{false} (clean).
The \texttt{"labels"} field records how sampled cells traverse the tree, including the cell identifier, predicted error label, and root-to-leaf \texttt{path}. These paths provide supervision for GNN node behavior, and make the resulting induced decision logic traceable.

\section{Algorithms} \label{sec:appendix_algorithms}
This section introduces the algorithms underlying \TreeTool{} and \ForestTool{}, including the decision tree construction and error prediction procedures in \TreeTool{}, as well as the overall algorithm of \ForestTool{}.

\noindent \textbf{Tree construction process.}
The overall workflow of decision tree construction process, summarized in Algorithm~\ref{algo:tree_refine}, follows a four-stage process. 
First, \emph{tree skeleton construction} creates an decision tree skeleton by defining node types and linking parent–child relationships according to the LLM-generated structure (lines 2-9). 
Next, \emph{label preparation} extracts supervision signals from the decision paths in labeled samples, assigning corresponding training labels to each GNN node (lines 10-14). 
Then, \emph{graph construction} represents the table as a bipartite graph connecting tuples and attributes, serving as the structural input for GNN training (lines 15-18). 
Finally, in \emph{GNN training and integration}, each GNN node is optimized and reattached to the tree, forming the final executable decision tree $\hat{\mathcal{T}}$ (lines 19-23). 

\begin{algorithm}[h]
\captionsetup{singlelinecheck=false}
\captionsetup{margin={0pt,5em}}
\caption{\parbox{\linewidth}{The algorithm for decision tree construction}}
\label{algo:tree_refine}
\LinesNumbered
\DontPrintSemicolon
\KwIn{Dirty table $\mathbf{D}$, LLM model $f_{LLM}$, samples $\mathcal{S}$}
\KwOut{Executable decision tree~$\hat{\mathcal{T}}$}

\textit{pmt} $\leftarrow$ prompt construction with $\mathcal{S}$\&$\mathbf{D}$; ~ $\mathcal{U}$, $\mathcal{\hat{S}} \leftarrow f_{llm}(\textit{pmt})$\;

\tcp*[l]{\textbf{Step 1: Tree skeleton construction}}
\ForEach{$u \in \mathcal{U}$}{
    \uIf{$u \in \mathcal{U}_{rule}$}{
        $\texttt{n} \leftarrow \{\texttt{type: rule},~u.\texttt{code},~u.\texttt{t\_child},~u.\texttt{f\_child}\}$\;

    }
    \uElseIf{$u \in \mathcal{U}_{gnn}$}{
        $\texttt{n} \leftarrow \{\texttt{type: gnn},~u.\texttt{t\_child},~u.\texttt{f\_child}\}$\;
    }
    \uElse{
        $\texttt{n} \leftarrow \{\texttt{type: leaf},~\texttt{value:}~\{0,1\}\}$\;
    }
    $\mathcal{T}_{\text{skeleton}} \gets \mathcal{T}_{\text{skeleton}} \cup \{\texttt{n}\}$\;
}

\tcp*[l]{\textbf{Step 2: Label preparation}}
\ForEach{$u_k \in \mathcal{U}_{gnn}$}{
    \ForEach{$s_{i,j} \in \hat{\mathcal{S}}$}{
        $\pi(i,j) \leftarrow \text{Flow}(\mathcal{T}_{\text{skeleton}}, s_{i,j})$\;
        \If{$(u_k, b_k) \in \pi(i,j)$}{
            $\mathcal{L}_{u_k} \leftarrow \mathcal{L}_{u_k} \cup \{(\mathbf{D}[i,j],~l_{i,j,k}=b_k)\}$\;
        }
    }
}
\tcp*[l]{\textbf{Step 3: Graph construction}}
\ForEach{$u_k \in \mathcal{U}_{gnn}$}{
    \ForEach{$i = 1...N, j=1..M$}{
        $\mathbf{h}_{e_{v_t, v_a}} \leftarrow \phi(\mathbf{D}[i,j]); \mathcal{E} \leftarrow \mathcal{E} \cup {e_{v_t, v_a}}$
    }
    $u_k.\texttt{graph} = (\mathcal{V}_t \cup \mathcal{V}_a, \mathcal{E}),~ u_k.\texttt{train\_labels} = \mathcal{L}_{u_k}$\;
}

\tcp*[l]{\textbf{Step 4: GNN training}}
\ForEach{$u_k \in \mathcal{U}_{gnn}$}{
    \ForEach{$i=1..N,\; j=1..M$}{
        $\hat{q}_{i,j,k} \leftarrow \text{MLP}\!\left(
\mathbf{h}_{v_{t}^i}^{(2)} \,\|\, \mathbf{h}_{v_{a}^j}^{(2)}
\right),$\;
    }
    Optimize $Loss_k \leftarrow -\!\!\sum_{(i,j)}
    [\,l_{i,j,k}\log\hat{q}_{i,j,k} + (1-l_{i,j,k})\log(1-\hat{q}_{i,j,k})\,]$\;
}

\Return $\hat{\mathcal{T}}$\;

\end{algorithm}

\begin{algorithm}[h]
\captionsetup{singlelinecheck=false}
\captionsetup{margin={0pt,5em}}
\caption{\TreeTool{} for error detection}
\label{algo:tree_prediction}
\LinesNumbered
\DontPrintSemicolon
\KwIn{Dirty table $\mathbf{D}$, LLM model $f_{LLM}$, samples $\mathcal{S}$}
\KwOut{Prediction matrix $\hat{\mathbf{Y}} \in \{0,1\}^{N\times M}$}
$\mathcal{\hat{T}} \leftarrow \text{Algorithm}~\ref{algo:tree_refine}(\mathbf{D}, f_{LLM}, \mathcal{S})$\;
\For{$i=1...N,\; j=1...M$}{
    $\text{node} \leftarrow \mathcal{\hat{T}}.{\text{root}}$\;
    \While{\text{n}.type $\neq$ leaf}{
        \uIf{{\text{node}.type $=$ rule}}{
            $\text{child} \leftarrow f_{\text{rule}}(D[i,j])$\;
        }
        \uIf{{\text{node}.type $=$ gnn}}{
            $\text{child}  \leftarrow f_{\text{gnn}}(i,j)$\;
        }
        $\text{node} \leftarrow \text{child} $\;
    }
    $\hat{\mathbf{Y}}[i,j] \leftarrow \text{node}.\text{value}$\;
}
\Return $\hat{\mathbf{Y}}$\;

\end{algorithm}

\noindent\textbf{Error prediction via decision tree.}
Once the decision tree $\hat{\mathcal{T}}$ is constructed, it can be directly applied to perform cell-level ED on a given table $\mathbf{D}$. 
As shown in Algorithm~\ref{algo:tree_prediction}, each cell is evaluated by following a root-to-leaf decision path: starting from the root (line~3), the model applies either a rule check or a GNN-based relational check at the current internal node. The Boolean outcome selects the next child node, and this process continues until a leaf is reached (lines~4-9).  
For each cell $\mathbf{D}[i,j]$, let $\pi(i,j)$ denote this root-to-leaf decision path, and let $u_K$ denote the leaf node at the end of the path; the value stored in $u_K$ becomes the final prediction for the cell (line~10).  
Iterating this procedure over all rows and columns (line~2) yields the binary prediction matrix $\hat{\mathbf{Y}} \in \{0,1\}^{N \times M}$ (line~11), where $\hat{\mathbf{Y}}[i,j] = 1$ indicates that $\mathbf{D}[i,j]$ is predicted as erroneous and $0$ otherwise.  

\noindent\textbf{Overall algorithm of \ForestTool{}.}
Algorithm~\ref{algo:forest_ed} summarizes the complete EM-based ensemble
consensus procedure in \ForestTool{}.  
Lines~1-2 perform sampling and partitioning, extracting a subset of informative rows from the original table.
Lines~3-4 generate $R$ tree-specific prediction matrices. 
Lines~5-17 implement the EM algorithm: the E-step (lines~7-8) refines the
posterior distribution about each cell's true label, while the M-step (lines~9-11)
updates each tree's reliability matrix and the overall prior distribution of error labels based on the inferred posteriors.  
This iterative refinement continues until the posterior distribution converges (line~12).  
Finally, it converts the posterior matrix into a binary consensus
prediction via MAP inference, yielding the final ED result (lines~13--15).

\begin{algorithm}[h]
\captionsetup{singlelinecheck=false}
\captionsetup{margin={0pt,5em}}
\caption{\ForestTool{}: EM-Based Ensemble Consensus for Error Detection}
\label{algo:forest_ed}
\LinesNumbered
\DontPrintSemicolon

\KwIn{Dirty table $\mathbf{D}$, LLM model $f_{LLM}$}
\KwOut{Final consensus matrix $\hat{\mathbf{Y}} \in \{0,1\}^{N\times M}$}

\tcp*[l]{\textbf{Step 1: Sampling and Partitioning}}
$\mathcal{}{S} \leftarrow \texttt{Sampling}(\mathbf{D})$\;
$\mathcal{S}^{(1)},\dots,\mathcal{S}^{(R)} \leftarrow \texttt{Partition}(\mathcal{S})$\;

\tcp*[l]{\textbf{Step 2: Tree-Level Predictions}}
\For{$r = 1 \ldots R$}{
    $\hat{\mathbf{Y}}^{(r)} \leftarrow \texttt{TreeED}(\mathbf{D}, f_{LLM}, \mathcal{S}^{(r)})$\;
}

\tcp*[l]{\textbf{Step 3: EM-Based Ensemble Consensus}}

Initialize posterior $\gamma_{i,j}(y)$ and confusion matrix 
$\Theta^{(r)}$\;

\Repeat{\text{convergence}}{

    \tcp*[l]{\textbf{E-step: update posteriors}}
   
    \For{each cell $(i,j)$; $y\in\{0,1\}$}{
            $\displaystyle
            \gamma_{i,j}(y)
            \leftarrow
            \frac{
                \eta_y
                \prod_{r=1}^R
                \theta^{(r)}_{\,y,\,\hat{\mathbf{Y}}^{(r)}_{i,j}}
            }{
                \sum_{y'\in\{0,1\}}
                \eta_{y'}
                \prod_{r=1}^R
                \theta^{(r)}_{\,y',\,\hat{\mathbf{Y}}^{(r)}_{i,j}}
            }
            $\;
    }

    \tcp*[l]{\textbf{M-step: update reliability matrices}}
    \For{$r = 1 \ldots R$; $y,\hat{y}\in\{0,1\}$}{
            $\displaystyle
            \Theta^{(r)}_{y,\hat{y}}
            \leftarrow
            \frac{
                \sum_{i,j: \,\hat{\mathbf{Y}}^{(r)}_{i,j} = \hat{y}}
                \gamma_{i,j}(y)
            }{
                \sum_{i,j} \gamma_{i,j}(y)
            }
            $\;
    }
    $\eta_y \leftarrow \frac{1}{NM}\sum_{i,j}\gamma_{i,j}(y)$\;
}

\tcp*[l]{\textbf{Final consensus prediction}}
\For{each cell $(i,j)$}{
    $\hat{\mathbf{Y}}_{i,j} \leftarrow \arg\max_{y\in\{0,1\}} \gamma_{i,j}(y)$\;
}

\Return $\hat{\mathbf{Y}}$\;
\end{algorithm}

\section{Discussion and Analysis}~\label{sec:appendix_analysis}

\noindent\textbf{Explainability.}  
\ForestTool{} provides explainability at both the tree and ensemble levels. At the tree level, each prediction is associated with a root-to-leaf decision path. At the ensemble level, the estimated confusion matrices quantify tree reliability, showing which induced reasoning patterns contribute more to the final consensus. Thus, analysts can trace both the local decision path and the tree-level evidence behind each predicted error.

\noindent\textbf{Time complexity analysis.}
Let the table contain $N \times M$ cells. For one \TreeTool{} tree, the LLM is invoked once with cost $C_{\text{LLM}}$, GNN training costs $\mathcal{O}(NMd)$, and tree inference over all cells costs $\mathcal{O}(NMk)$, where $d$ is the embedding dimension and $k$ is the tree depth. Thus, the time complexity of \TreeTool{} is
$\mathcal{O}(NM(d+k)+C_{\text{LLM}})$.
With $R$ induced trees, \ForestTool{} requires
$\mathcal{O}(R(NM(d+k)+C_{\text{LLM}}))$ before consensus. The EM consensus further costs $\mathcal{O}(INMR)$ over $I$ iterations, yielding an overall complexity of
$\mathcal{O}(R(NM(d+k)+C_{\text{LLM}})+INMR)$.

\noindent \textbf{Proof of Lemma~\ref{lem:uncertainty_sampling}.}
~\label{proof:uncertainty_sampling}

\begin{proof}
For a sampled tuple $s$, the information gain is
\[
I(s) = H(\mathcal{V}) - H(\mathcal{V}\mid s),
\]
where $\mathcal{V}$ is the version space.
Let $\mathcal{N}(\mathcal{B})$ denote a small neighborhood around the true boundary $\mathcal{B}$.
Due to error sparsity, we have
\[
\Pr(s \in \mathcal{N}(\mathcal{B})) \ll 1.
\]

Under random sampling, most tuples satisfy $s \notin \mathcal{N}(\mathcal{B})$, for which the posterior label distribution is highly concentrated, implying
\[
H(\mathcal{V}\mid s) \approx H(\mathcal{V}) \quad \Rightarrow \quad I(s) \approx 0.
\]
Thus, the expected information gain satisfies
\[
\mathbb{E}[I(\mathcal{S}_r)] 
= \mathbb{E}_{s \sim \mathcal{S}_r}[I(s)]
\approx \int_{s \notin \mathcal{N}(\mathcal{B})} 0 \, ds
\approx 0.
\]
In contrast, uncertainty-based sampling preferentially selects tuples with high predictive variance,
\[
\operatorname{Var}(y \mid s) \text{ is large},
\]
which concentrates probability mass on $s \in \mathcal{N}(\mathcal{B})$~\cite{krishnan2016activeclean}.
For such tuples, multiple hypotheses in $\mathcal{V}$ disagree, yielding
\[
H(\mathcal{V}\mid s) < H(\mathcal{V}) \quad \Rightarrow \quad I(s) > 0.
\]
Therefore,
\[
\mathbb{E}[I(\mathcal{S}_u)] 
= \mathbb{E}_{s \sim \mathcal{S}_u}[I(s)] 
> 0.
\]
Combining the above results yields
\[
\mathbb{E}[I(\mathcal{S}_u)] > \mathbb{E}[I(\mathcal{S}_r)],
\]
which completes the proof.
\end{proof}

\noindent\textbf{Proof of Lemma~\ref{lemma1}.} ~\label{proof}
\begin{proof}

Using Jensen’s inequality,
\[
\log p(\hat{\mathbf{Y}}\mid \Theta)
\ge
\mathcal{F}(o,\Theta),
\]
so $\mathcal{F}$ is a lower bound on the marginal log-likelihood.
Fixing $\Theta$, maximizing $\mathcal{F}$ over $o$ yields
\[
o_{i,j}(y)
=
p(\mathbf{Y}_{i,j}=y\mid\hat{\mathbf{Y}},\Theta),
\]
which uniquely maximizes the bound and increases $\mathcal{F}$ unless 
already optimal.  
Fixing $o$, maximizing $\mathcal{F}$ over each $\Theta^{(r)}$ under the 
constraint $\sum_{\hat y}\theta^{(r)}_{y,\hat y}=1$ gives
\[
\theta^{(r)}_{y,\hat y}
=
\frac{
\sum_{i,j:\hat{\mathbf{Y}}^{(r)}_{i,j}=\hat y}
o_{i,j}(y)
}{
\sum_{i,j} o_{i,j}(y)
}.
\]
Thus each E-step and M-step maximizes the ELBO, ensuring
\[
\mathcal{F}(o^{(t+1)},\Theta^{(t+1)})
\ge
\mathcal{F}(o^{(t)},\Theta^{(t)}),
\]
which implies a monotonic increase in 
$\log p(\hat{\mathbf{Y}}\mid\Theta)$.  
Since $o$ determines posterior consensus and $\Theta$ encodes tree
reliability, improving the ELBO directly improves ensemble consensus.

\end{proof}

\vspace{-2mm}

\section{Additional Experiments}
~\label{sec:appendix_experiment}

\begin{figure}[t]
    \centering

    \begin{subfigure}{0.225\textwidth}
        \centering
        \includegraphics[width=\linewidth]{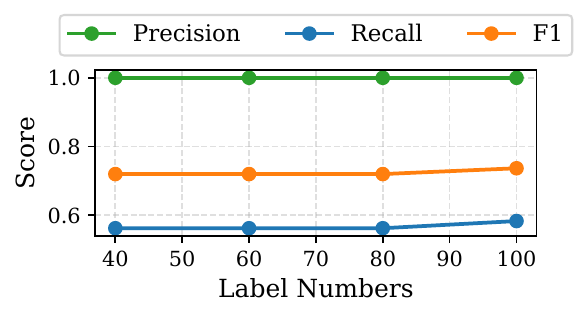}
        \caption{Flights}
    \end{subfigure}
    \hfill
    \begin{subfigure}{0.225\textwidth}
        \centering
        \includegraphics[width=\linewidth]{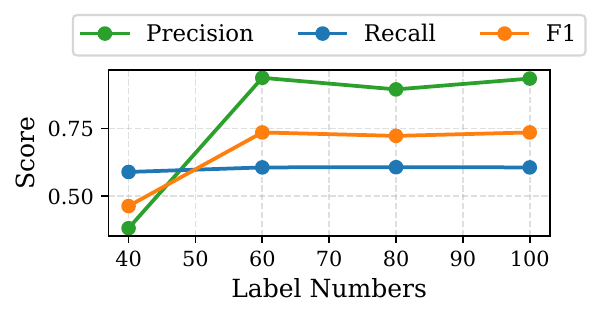}
        \caption{Movies}
    \end{subfigure}

    \caption{Effect of varying the number of labeled records on model performance across different datasets.}
    \label{fig:exp4_label_rate}
      \vspace{-1mm}

\end{figure}

\noindent\textbf{Exp-9: Varying numbers of labels.}
We vary the number of labeled records from 40 to 100 while keeping other settings fixed. As shown in Figure~\ref{fig:exp4_label_rate}, additional labels generally improve accuracy and stabilize performance, though the gain varies across datasets. \textit{Flights} remains stable even with 40 labels, while \textit{Movies} improves substantially from 40 to 60 labels and then stabilizes. Based on these results, we use 100 labeled tuples as the upper bound to ensure stable performance without excessive annotation.

\noindent\textbf{Exp-10: Varying LLMs.}
Table~\ref{tab:varying_llms} shows that \ForestTool{} maintains more stable performance across different LLM backbones. By contrast, \ZeroED{} is highly sensitive to backbone selection, with an average F1-score drop of about 25.5\% between its best and worst-performing backbones. In contrast, ForestED exhibits much smaller fluctuations (average F1-score spread of only 5.25\% across backbones) and consistently outperforms the corresponding ZeroED variant under every LLM. This demonstrates that ForestED’s multi-tree design and EM-based consensus effectively buffer backbone-induced noise, providing more stable and reliable performance regardless of the underlying LLM.

\begin{table}[t]
\centering
\small
\caption{Performance comparison of \ForestTool{} and \ZeroED{} using different LLM models on different datasets.}
\label{tab:varying_llms}
\setlength{\tabcolsep}{3pt}
\begin{tabular}{
|p{2.0cm}<{\centering}|
p{0.8cm}<{\centering}p{0.8cm}<{\centering}p{0.8cm}<{\centering}|
p{0.8cm}<{\centering}p{0.8cm}<{\centering}p{0.8cm}<{\centering}|
}
\hline
\multirow{2}{*}{\textbf{LLM Models}} &
\multicolumn{3}{c|}{\textbf{Rayyan}} &
\multicolumn{3}{c|}{\textbf{Flights}} \\ 
\cline{2-7}
 & Prec & Rec & F1 & Prec & Rec & F1 \\ 
\hline
\multicolumn{7}{|c|}{\cellcolor{lightgray}\textbf{ForestED}} \\ \hline
GPT-5                      & \textbf{0.998} & 0.544 & 0.704  & \textbf{1.000} & \textbf{0.562} & \textbf{0.720} \\ 
DeepSeek-V3                & 0.992 & \textbf{0.575} & \textbf{0.728} & 0.993 & 0.531 & 0.692 \\ 
Qwen3                      & \textbf{0.998}    & 0.543    & 0.704    & \textbf{1.000} & 0.470 & 0.639 \\ 
\hline
\multicolumn{7}{|c|}{\cellcolor{lightgray}\textbf{ZeroED}} \\ \hline
GPT-5                      & \textbf{0.751} & 0.631 & \textbf{0.686} & \textbf{0.493} & \textbf{0.720} & \textbf{0.585} \\ 
DeepSeek-V3                & 0.270 & 0.646 & 0.375 & 0.174 & 0.603 & 0.387 \\ 
Qwen3                      & 0.365 & \textbf{0.698} & 0.479 & 0.430 & 0.687 & 0.529 \\ 
\hline
\end{tabular}
\end{table}

\noindent\textbf{Exp-11: Varying tree numbers.}  
This experiment studies the effect of ensemble size on \ForestTool{}. As shown in Table~\ref{tab:exp8_tree_numbers}, small ensembles with 2 or 5 trees provide limited coverage of diverse error patterns, leading to suboptimal performance. Increasing the number to 10 achieves the best overall results, suggesting a good balance between diversity and stability. Further increasing it to 20 slightly degrades performance, likely due to overfitting and noise propagation. Configurations marked by ``--'' indicate that the LLM failed to return valid responses within the default time window. Overall, \ForestTool{} performs most stably with around 10 trees per dataset.

\begin{table}[t]
\centering
\caption{Performance of \ForestTool{} under different numbers of trees. 
“--” indicates that the LLM timed out.
}
\label{tab:exp8_tree_numbers}

\resizebox{0.48\textwidth}{!}{
\begin{tabular}{|c|ccc|ccc|}
\hline
\multirow{2}{*}{\makecell{Tree \\ Number}} &
\multicolumn{3}{c|}{\textbf{Flights}} &
\multicolumn{3}{c|}{\textbf{Movies}} \\ \cline{2-7}
 & Prec & Rec & F1 & Prec & Rec & F1 \\ \hline
2  & 1.000 & 0.470 & 0.639 & -- & -- & -- \\
5  & 1.000 & 0.556 & 0.715 & -- & -- & -- \\
10 & \textbf{1.000} & \textbf{0.562} & \textbf{0.720} & \textbf{0.934} & \textbf{0.606} & \textbf{0.735} \\
20 & 1.000 & 0.562 & 0.720 & 0.182 & 0.698 & 0.289 \\ \hline
\end{tabular}
}
\end{table}

\begin{table}[t]
\centering
\caption{Performance of \ForestTool{} under different tree depths. }
\label{tab:exp9_tree_depths}

\resizebox{0.48\textwidth}{!}{
\begin{tabular}{|c|ccc|ccc|}
\hline
\multirow{2}{*}{\makecell{Tree \\ Depth}} &
\multicolumn{3}{c|}{\textbf{Hospital}} &
\multicolumn{3}{c|}{\textbf{Beers}} \\ \cline{2-7}
 & Prec & Rec & F1 & Prec & Rec & F1 \\ \hline
2$\sim$6  & 0.769 & 0.159 & 0.263 & 0.917 & 0.951 & 0.934 \\
4$\sim$8  & \textbf{0.844} & 0.685 & \textbf{0.756} & \textbf{0.974} & 0.922 & \textbf{0.947} \\
6$\sim$10 & 0.482 & \textbf{0.700} & 0.571 & 0.881 & \textbf{0.953} & 0.916 \\ \hline
\end{tabular}
}
\end{table}

\noindent\textbf{Exp-12: Varying tree depths.}  
We further study how the depth of individual decision trees influences the detection performance of \ForestTool{}.  
Table~\ref{tab:exp9_tree_depths} reports results across two representative datasets.  
Shallow trees (depth 2–6) show limited capacity, leading to underfitting and low recall, while overly deep trees (depth 6–10) tend to overfit noisy records.
The optimal performance is achieved when the tree depth ranges from 4–8, balancing readability and expressiveness, and yielding the highest F1-scores across both datasets.

\end{document}